\documentclass{article}
\usepackage[accepted]{icml2020}
\usepackage[dvipsnames]{xcolor}
\usepackage{microtype}
\usepackage{graphicx,epsfig}
\usepackage{booktabs} 

\usepackage[toc,page,header]{appendix}
\usepackage{minitoc}

\usepackage{subcaption}
\usepackage{epstopdf}
\usepackage{amsthm}
\usepackage[colorinlistoftodos]{todonotes} 
\usepackage{hyperref}

\usepackage{breakurl}
\usepackage{float}
\usepackage{xcolor}
\usepackage{hhline}
\usepackage{multirow}

\usepackage{amsmath}
\usepackage{amssymb}
\usepackage{amsfonts}

\newcommand{\reals}{\mathbb{R}}
\newcommand{\real}{\mathbb{R}}
\newcommand{\dual}{\vec{v}}
\newcommand{\dualmat}{\vec{V}}

\newcommand{\rectset}{\mathcal{Q}_\data}
\newcommand{\relu}[1]{\big( #1 \big)_+}
\newcommand{\ball}{\mathcal{B}}

\newcommand{\sign}{\text{sign}}
\newcommand{\D}{\mbox{Diag}}

\let\vec\notbold
\let\boldsymbol\notbold

\newtheorem{remark}{Remark}[section]

\theoremstyle{definition}

\DeclareMathOperator{\trace}{trace}

\icmltitlerunning{Neural Networks are Convex Regularizers}

\begin{document}
\doparttoc 
\faketableofcontents 

\twocolumn[
\icmltitle{Neural Networks are Convex Regularizers: Exact Polynomial-time Convex Optimization Formulations for Two-layer Networks}

\icmlsetsymbol{equal}{*}

\begin{icmlauthorlist}
\icmlauthor{Mert Pilanci}{to}
\icmlauthor{Tolga Ergen}{to}

\end{icmlauthorlist}

\icmlaffiliation{to}{Department of Electrical Engineering, Stanford University, CA, USA}

\icmlcorrespondingauthor{Mert Pilanci}{pilanci@stanford.edu}

\icmlcorrespondingauthor{Tolga Ergen}{ergen@stanford.edu}

\vskip 0.2in
]


\printAffiliationsAndNotice{} 

\newtheorem{theorem}{Theorem}
\newtheorem{corollary}{Corollary}[theorem]
\newtheorem{lemma}[theorem]{Lemma}

\begin{abstract}
We develop exact representations of training two-layer neural networks with rectified linear units (ReLUs) in terms of a single convex program with number of variables polynomial in the number of training samples and the number of hidden neurons. Our theory utilizes semi-infinite duality and minimum norm regularization. We show that ReLU networks trained with standard weight decay are equivalent to block $\ell_1$ penalized convex models. Moreover, we show that certain standard convolutional linear networks are equivalent semi-definite programs which can be simplified to $\ell_1$ regularized linear models in a polynomial sized discrete Fourier feature space.
\end{abstract}
\newcommand{\data}{X}
\newcommand{\datavec}{x}
\newcommand{\labelvec}{y}
\newcommand{\firstw}{u}
\newcommand{\secondw}{\alpha}
\newcommand{\act}{\phi}
\newcommand{\bias}{b}
\newcommand{\weightone}{w}

\newcommand{\weightscalar}{w}
\newcommand{\weight}{\vec{w}}
\newcommand{\weightmat}{\vec{W}}
\vspace{-0.5cm}
\section{Introduction}

In this paper, we introduce a finite dimensional, polynomial-size convex program that globally solves the training problem for two-layer neural networks with rectified linear unit (ReLU) activation functions. The key to our analysis is a generic convex duality method we introduce, and is of independent interest for other non-convex problems. We further prove that strong duality holds in a variety of architectures.
\subsection{Related work and overview}
Convex neural network training was considered in the literature \cite{bengio2006convex,bach2017breaking}. However, convexity arguments in the existing work are restricted to infinite width networks, where infinite dimensional optimization problems need to be solved. In fact, adding even a single neuron to the model requires the solution of a non-convex problem where no efficient algorithm is known \cite{bach2017breaking}. In this work, we develop a novel duality theory and introduce polynomial-time finite dimensional convex programs, which are exact and computationally tractable.

Several recent studies considered over-parameterized neural networks, where the width approaches infinity by leveraging connections to kernel methods, and showed that randomly initialized gradient descent can fit all the training samples \cite{jacot2018neural,du2018gradient,allen-zhu19a}. However, in this \emph{kernel regime}, the analysis shows that almost no hidden neurons move from their initial values to actively learn useful features \cite{lazy_training_bach}. Experiments also confirm that the kernel approximation as the width tends to infinity is unable to fully explain the success of non-convex neural network models \cite{arora2019exact}. On the contrary, our work precisely characterizes the mechanism behind extraordinary modeling capabilities of neural networks for any finite number of hidden neurons. We prove that networks with ReLU are identical to \emph{convex regularization} methods in a finite higher dimensional space.

Consider a two-layer network $f:\mathbb{R}^d \rightarrow \mathbb{R}$ with $m$ neurons
\begin{align}
    f(\datavec) =\sum_{j=1}^m  \act(x^T \firstw_j) \secondw_j\,, \label{eq:2layer_network}
\end{align}
where $\firstw_j \in \real^d$ and $\secondw_j \in \real$ are the weights for  hidden and output layers, respectively, and $\phi(t)=(t)_+:=\max(t,0)$ is the ReLU activation. We extend the definition of scalar functions to vectors/matrices entry-wise. We use $\ball_p$ to denote the unit $\ell_p$ ball in $\real^d$. We denote the set of integers from $1$ to $n$ as $[n]$. We also use $\sigma$ to denote singular values.
%

In order to keep the notation simple and clearly convey the main idea, we will restrict our attention to two-layer  ReLU networks with scalar output trained with squared loss. All of our results immediately extend to vector outputs, tensor inputs, arbitrary convex classification and regression loss functions, and other network architectures (see Appendix).

	\begin{figure*}[t]
	\centering
	\captionsetup[subfigure]{oneside}
		\begin{subfigure}[t]{0.32\textwidth}
		\centering
		\includegraphics[width=0.95\textwidth, height=0.9\textwidth]{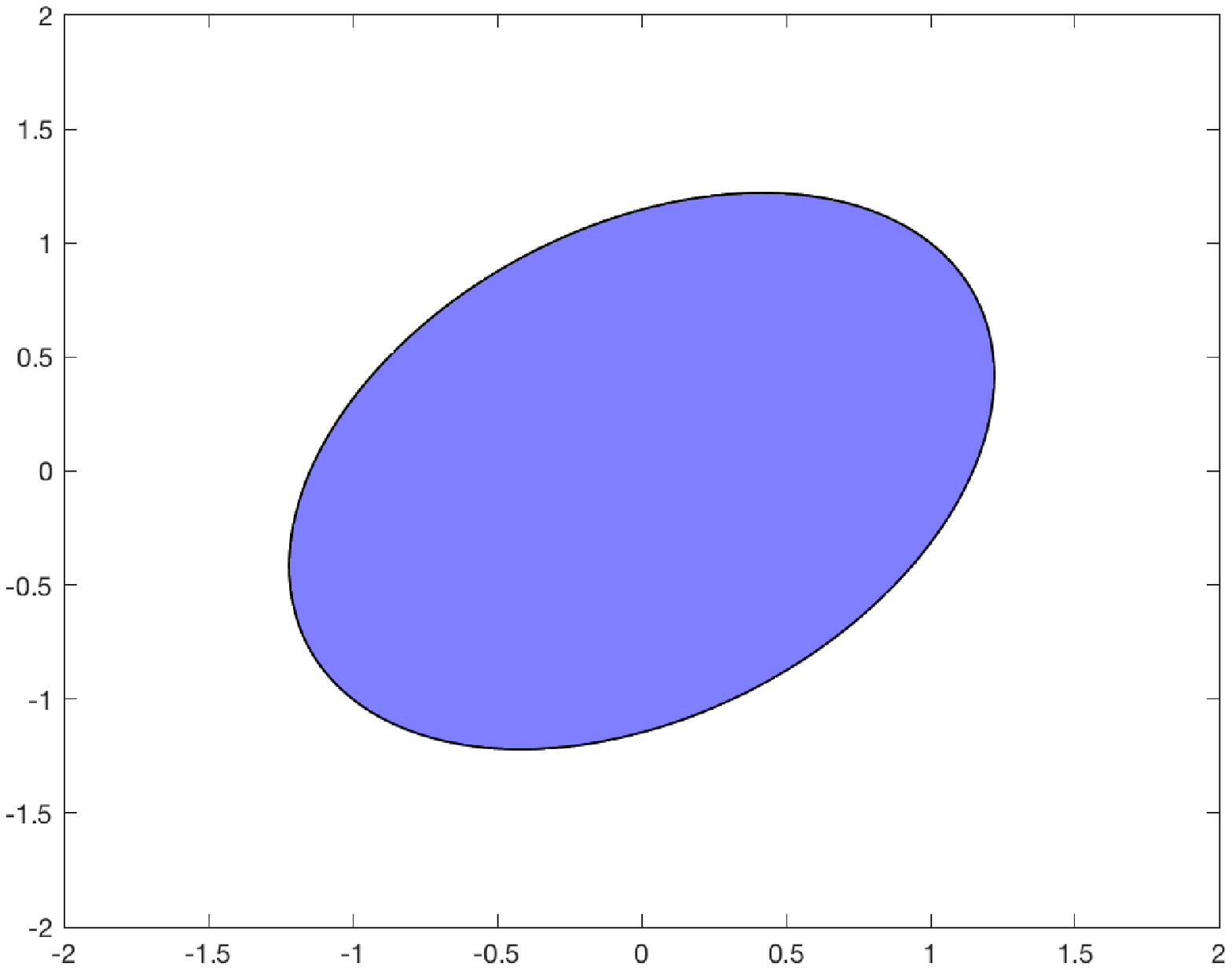}
		\caption{Ellipsoidal set: $\{\data u : u \in \mathbb{R}^d, \|  u\|_2 \leq 1\}$\centering} \label{fig:spike2_ellipsoid}
	\end{subfigure} \hspace*{\fill}
		\begin{subfigure}[t]{0.32\textwidth}
		\centering
		\includegraphics[width=0.95\textwidth, height=0.9\textwidth]{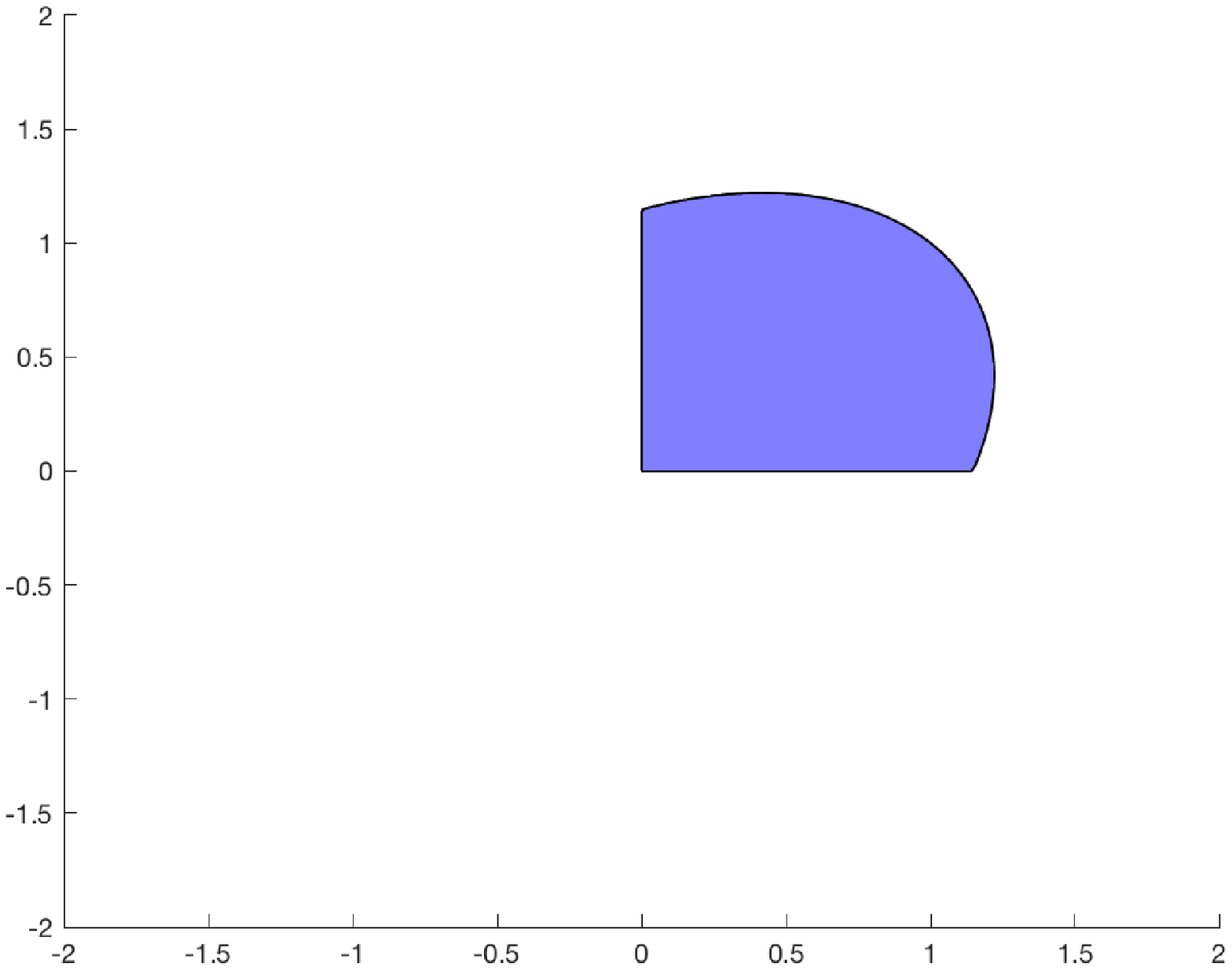}
		\caption{Rectified ellipsoidal set $\rectset$: $\big \{ \relu{ \data  u } : u \in \mathbb{R}^d, \|  u\|_2 \leq 1   \big\}$\centering} \label{fig:spike2_recellipsoid}
	\end{subfigure} \hspace*{\fill}
			\begin{subfigure}[t]{0.32\textwidth}
		\centering
		\includegraphics[width=0.95\textwidth, height=0.9\textwidth]{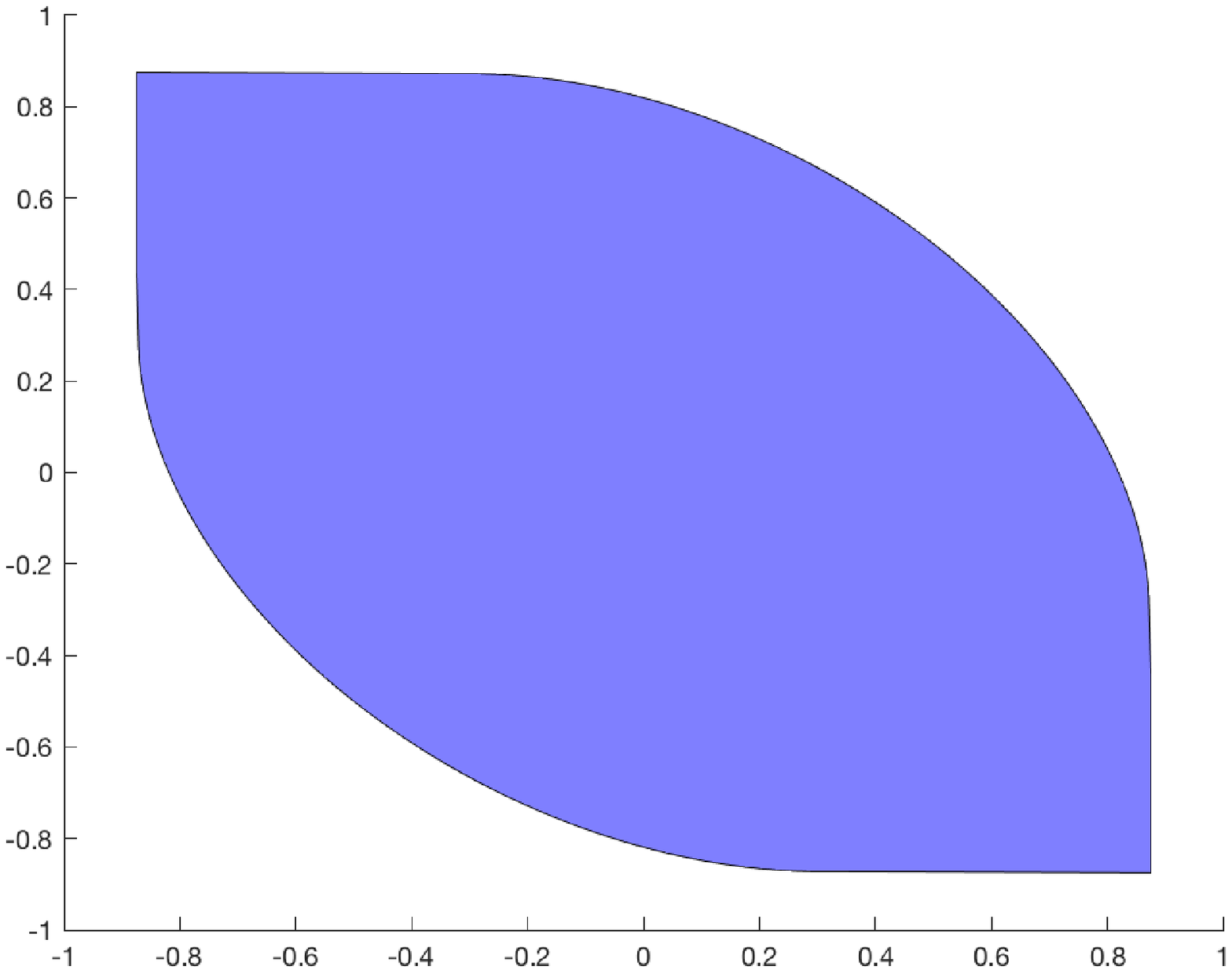}
		\caption{Polar set $(\rectset\cup-\rectset)^\circ$: $\{\dual : \vert\, \dual^T \weight \vert \le 1\,, \forall \weight\in \rectset\}$\centering} \label{fig:spike2_polar}
	\end{subfigure} \hspace*{\fill}
	\medskip
	\caption{Sets involved in the construction of the Neural Gauge. Ellipsoidal set, rectified ellipsoid $\rectset$ and the polar of $\rectset\cup-\rectset$.}
\label{fig:spike2}
\vskip -0.2in
	\end{figure*}
%
Given a data matrix $X \in \real^{n \times d}$, a label vector $y \in \real^n$, and a regularization parameter $\beta>0$, consider minimizing the squared loss objective and squared $\ell_2$-norm of all parameters
\begin{align} \label{eq:2layer_regularized_cost}
& p^*:=\hspace{-0.3cm}\min_{\{\secondw_j,\firstw_j\}_{j=1}^m} \frac{1}{2}\Big\| \sum_{j=1}^m (\data \firstw_j )_+ \secondw_{j} -\labelvec \Big\|_2^2 \\ \nonumber & \hspace{4.5cm}+\frac{\beta}{2} \sum_{j=1}^m (\|\firstw_j\|_2^2+\secondw_j^2)\,. 
\end{align}
The above objective is highly non-convex due to non-linear ReLU activations and product between hidden and outer layer weights.
The best known algorithm for globally minimizing the above objective is a brute-force search over all possible  piece-wise linear regions of ReLU activations of $m$ neurons and output layer sign patterns \cite{arora2018understanding}. This algorithm has complexity $O(2^m n^{dm})$ (see Theorem 4.1 in \cite{arora2018understanding}). In fact, known algorithms for approximately learning $m$ hidden neuron ReLU networks have complexity $O(2^{\sqrt{m}})$ (see Theorem 5 of \cite{goel17a}) due to similar combinatorial explosion with $m$.

\section{Convex Duality for Two-layer Networks}
Now we introduce our main technical tool for deriving convex representations of the non-convex objective function \eqref{eq:2layer_regularized_cost}. We start with the $\ell_1$ penalized representation, which is equivalent to \eqref{eq:2layer_regularized_cost} (see Appendix \ref{sec:equivalence_appendix}),
\begin{align}
     p^*=\min_{\substack{\|u_j\|_2\le 1 \\ \forall j\in [m]}} \min_{\{\secondw_j\}_{j=1}^m} \frac{1}{2}\Big\| \sum_{j=1}^m (\data \firstw_j )_+ \secondw_j -\labelvec \Big\|_2^2 +\beta \sum_{j=1}^m \vert \secondw_j \vert\,.
    \label{eq:2layer_regularized_cost_l1}
\end{align}

Replacing the inner minimization problem with its convex dual, we obtain (see Appendix \ref{sec:twolayer_dualform_appendix})
\begin{align*}
     p^*=\min_{\substack{\|u_j\|_2\le 1 \\ \forall j\in [m]}} \max_{ \substack{v \in \real^n \,\mbox{\scriptsize s.t.} \\ \vert v^T(\data u_j)_+\vert \le \beta,\,\forall j\in[m]  }} -\frac{1}{2} \|y-v\|_2^2 + \frac{1}{2}\|y\|_2^2\,.
\end{align*}
Interchanging the order of $\min$ and $\max$, we obtain the lower-bound $d^*$ via weak duality
\begin{align}
     p^*\ge d^*:= \hspace{-0.3cm} \max_{ \substack{v \in \real^n \,\mbox{\scriptsize s.t.} \\ \vert v^T(\data u)_+\vert \le \beta\, \forall u \in \ball_2  }} -\frac{1}{2} \|y-v\|_2^2 + \frac{1}{2}\|y\|_2^2\,.
    \label{eq:2layer_regularized_cost_innerdual}
\end{align}
The above problem is a convex \emph{semi-infinite} optimization problem with $n$ variables and infinitely many constraints. We will show that strong duality holds, i.e., $p^*=d^*$, as long as the number of hidden neurons $m$ satisfies $m\ge m^*$ for some $m^*\in \mathbb{N}$, $1\le m^* \le n$, which will be defined in the sequel. As it will be shown, $m^*$ can be smaller than $n$.  
 The dual of the dual program \eqref{eq:2layer_regularized_cost_innerdual} can be derived using standard semi-infinite programming theory \cite{semiinfinite_goberna}, and corresponds to the bi-dual of the non-convex problem \eqref{eq:2layer_regularized_cost}.

 Now we briefly introduce basic properties of signed measures that are necessary to state the dual of \eqref{eq:2layer_regularized_cost_innerdual} and refer the reader to \cite{rosset2007,bach2017breaking} for further details. Consider an arbitrary measurable input space $\mathcal{X}$ with a set of continuous basis functions $\phi_{\vec{u}}:\mathcal{X}\rightarrow \reals$ parameterized by $\vec{u}\in \ball_2$. We then consider real-valued Radon measures equipped with the uniform norm \cite{Rudin}. For a signed Radon measure $\boldsymbol{\mu}$, we can define an infinite width neural network output for the input $\vec{x}\in\mathcal{X}$ as $f(\vec{x}) = \int_{\vec{u}\in \ball_2} \phi_\vec{u}(\vec{x}) d\boldsymbol{\mu}(\vec{u})$\,. The total variation norm of the signed measure $\boldsymbol{\mu}$ is defined as the supremum of $\int_{\vec{u}\in\ball_2}q(\vec{u})d\boldsymbol{\mu}(\vec{u})$ over all continuous functions $q(\vec{u})$ that satisfy $|q(\vec{u})|\le 1$. Consider the ReLU basis functions $\phi_{\vec{u}}(\vec{x})=\relu{\vec{x}^T\vec{u}}$. We may express networks with finitely many neurons as in \eqref{eq:2layer_network} by
\begin{align*}
    f(\vec{x}) = \sum_{j=1}^m \phi_{\vec{u}_j}(\vec{x}) \secondw_j\,,
\end{align*}
which corresponds to $\boldsymbol{\mu} = \sum_{j=1}^m \secondw_j \delta(\vec{u}-\vec{u}_j)$ where $\delta$ is the Dirac delta measure. And the total variation norm $\|\boldsymbol{\mu}\|_{TV}$ of $\boldsymbol{\mu}$ reduces to the $\ell_1$ norm $\|\secondw\|_1$.

We state the dual of \eqref{eq:2layer_regularized_cost_innerdual} (see Section 2 of \cite{shapiro2009semi} and Section 8.6 of \cite{semiinfinite_goberna}) as follows
\begin{align}
    d^*\le p_{\infty}^*=\min_{\mu} \frac{1}{2} \left\| \int_{\vec{u}\in \ball_2} \relu{\data\vec{u}}d\boldsymbol{\mu}(\vec{u}) - \vec{y} \right\|_2^2 + \beta\, \|\boldsymbol{\mu}\|_{TV}. \label{eq:infdim}
\end{align}
%
Furthermore, an application of Caratheodory's theorem shows that the infinite dimensional bi-dual \eqref{eq:infdim} always has a solution that is supported with $m^*$ Dirac deltas, where $m^*\le n+1$ \cite{rosset2007}. Therefore we have
\begin{align*}
     p_{\infty}*&=\min_{\substack{ \|u_j\|_2\le 1 \\ \{\secondw_j,u_j\}_{j=1}^{m^*}}} \frac{1}{2}\Big\| \sum_{j=1}^{m^*} (\data \firstw_j )_+ \secondw_j -\labelvec \Big\|_2^2 +\beta \sum_{j=1}^{m^*} \vert \secondw_j \vert\,, 
    \\
    &=p^*\,,
\end{align*}
as long as $m\ge m^*$. We show that strong duality holds, i.e., $d^*=p^*$ in Appendix \ref{appendix_semi_infinite_duality} and \ref{appendix_semi_infinite_duality2}. In the sequel, we illustrate how $m^*$ can be determined via a finite-dimensional parameterization of \eqref{eq:2layer_regularized_cost_innerdual} and its dual.
\subsection{A geometric insight: Neural Gauge Function}
An interesting geometric insight can be provided in the weakly regularized case where $\beta \rightarrow 0$. In this case, minimizers of \eqref{eq:2layer_regularized_cost_l1} and hence \eqref{eq:2layer_regularized_cost} approach minimum norm interpolation $p_{\beta\rightarrow 0}^*:=\lim_{\beta\rightarrow 0} \beta^{-1}p^*$ given by
\begin{align}
    p^*_{\beta\rightarrow 0}=&\min_{\{u_j,\alpha_j\}_{j=1}^m} \sum_{j=1}^m \vert \alpha_j \vert\\
    &\mbox{s.t. } \sum_{j=1}^m(Xu_j)_+\alpha_j=y,\, \|u_j\|_2\le 1\,\forall j.\textbf{} \nonumber
\end{align}
We show that $p^*_{\beta\rightarrow 0}$ is the gauge function of the convex hull of $\rectset \cup -\rectset$ where $\rectset:=\{ (\data u)_+\,:\,u\in \ball_2 \}$ (see Appendix \ref{sec:gauge_appendix}), i.e.,
\begin{align*}
    p^*_{\beta\rightarrow 0} = &\inf_{t:t\ge 0} \,\, t \mbox{ s.t. }
     y \in t \,\mbox{Conv}\{{\rectset \cup -\rectset\}}\,,
\end{align*}
which we call \emph{Neural Gauge} due to the connection to the minimum norm interpolation problem. Using classical polar gauge duality (see e.g. \cite{Rockafellar}, it holds that
\begin{align}\label{eq:supportfun}
    p^*_{\beta\rightarrow 0}= &\max y^T z  \mbox{ s.t. } z \in (\rectset \cup -\rectset)^\circ \,,
\end{align}
where $(\rectset \cup -\rectset)^\circ$ is the polar of the set $\rectset \cup -\rectset$. Therefore, evaluating the support function of this polar set is equivalent to solving the neural gauge problem, i.e., minimum norm interpolation  $p^*_{\beta\rightarrow 0}$. These sets are illustrated in Figure \ref{fig:spike2}. Note that the polar set $(\rectset \cup -\rectset)^\circ$ is always convex (see Figure \ref{fig:spike2}c),  which also appears in the dual problem $\eqref{eq:2layer_regularized_cost_innerdual}$ as a constraint. In particular, $\lim_{\beta\rightarrow 0} \beta^{-1}d^*$ is equal to the support function. Our finite dimensional convex program leverages the convexity and an efficient description of this set as we discuss next.

\section{An Exact Finite Dimensional Convex Program}
%
Consider diagonal matrices $\D(1[\data u\ge 0])$ where $u\in\real^d$ is arbitrary and $1[\data u\ge 0] \in \{0,1\}^n$ is an indicator vector with Boolean elements $[1[\datavec_1^Tu\ge 0] ,\dots,1[\datavec_n^Tu\ge 0]]$. Let us enumerate all such distinct diagonal matrices that can be obtained for all possible $u\in \real^d$, and denote them as $D_1,...,D_P$. $P$ is the number of regions in a partition of $\real^d$ by hyperplanes passing through the origin, and are perpendicular to the rows of $\data$. It is well known that $$P\le 2 \sum_{k=0}^{r-1} {n-1 \choose k}\le 2r\Big(\frac{e(n-1)}{r}\Big)^r \,$$ for $r\le n$, where $r:=\mbox{rank}(\data)$ \cite{ojha2000enumeration,stanley2004introduction,winder1966partitions,cover1965geometrical} (see Appendix \ref{sec:hyperplane_arrangments_appendix}).

Consider the finite dimensional convex problem
\begin{align}
&\min_{\{v_i,w_i\}_{i=1}^P}\, { \frac{1}{2}}\Big \| \sum_{i=1}^P D_iX(v_i-w_i) - y \Big\|_2^2\nonumber \\  &\hspace{4cm}+\beta\sum_{i=1}^P  \left(\|v_i\|_2+\|w_i\|_2\right) \nonumber  \\
&\mbox{ s.t. } 
 (2D_i-I_n)Xv_i\ge0,~ (2D_i-I_n) Xw_i\ge 0,\, \forall i.\label{eq:twolayerconvexprogram}   
\end{align}
%
%
\begin{theorem}
\label{thm:mainconvex}
The convex program \eqref{eq:twolayerconvexprogram} and the non-convex problem \eqref{eq:2layer_regularized_cost} where $m\ge m^*$ have identical optimal values\footnote{$m^*$ is defined as the number of Dirac deltas in the optimal solution to \eqref{eq:infdim}. If the optimum is not unique, we may pick the minimum cardinality solution.}. Moreover, an optimal solution to \eqref{eq:2layer_regularized_cost} with $m^*$ neurons can be constructed from an optimal solution to \eqref{eq:twolayerconvexprogram} as follows
\begin{align*}
    &(u^*_{j_{1i}},\alpha^*_{j_{1i}}) =  \left(\frac{v^*_i}{\sqrt{\|v^*_i\|_2}}, \sqrt{\|v^*_i\|_2}\right) \,\quad \quad \mbox{  if  } \quad v_i^*\neq 0  \\
    &(u^*_{j_{2i}},\alpha^*_{j_{2i}}) =   \left(\frac{w^*_i}{\sqrt{\|w^*_i\|_2}}, -\sqrt{\|w^*_i\|_2}\right) \quad \mbox{  if  } \quad w_i^*\neq 0\,, 
\end{align*}
where $\{v_i^*,w_i^*\}_{i=1}^P$ are the optimal solutions to \eqref{eq:twolayerconvexprogram}. Thus, we have $m^*=\sum_{\tiny i: v_i^*\neq 0}^P 1+\sum_{\tiny i:  w^*_i\neq 0}^P 1$.
\end{theorem}
\begin{remark}
Note that optimal solutions of \eqref{eq:twolayerconvexprogram} may not be unique. As an example,  merging positively colinear neurons does not change the objective value. Particularly, if there exist positively colinear solutions such as $v_{i_1}^*=c_1 v_i$ and $v_{i_2}^*=c_2 v_i$, where $v_i \in \mathbb{R}^d$ and $c_1,c_2 \in \mathbb{R}_+$, then merging these solutions as $v_i^*=(c_1+c_2)v_i$ yields the same objective. 
\end{remark}
\begin{remark}
Theorem \ref{thm:mainconvex} shows that two-layer ReLU networks with $m$ hidden neurons can be globally optimized via the second order cone program \eqref{eq:twolayerconvexprogram} with $2dP$ variables and $2nP$ linear inequalities where  $P=2r\Big(\frac{e(n-1)}{r}\Big)^r$, and $r=\mbox{rank}(\data)$. The computational complexity is at most $O\Big(d^3r^3\big(\frac{n}{r}\big)^{3r}\Big)$ using standard interior-point solvers. For fixed rank $r$ (or dimension $d$), the complexity is polynomial in $n$ and $m$, which is an exponential improvement over the state of the art \cite{arora2018understanding,bienstock2018principled}. 
Note that $d$ is a small number that corresponds to the filter size in CNNs as we illustrate in the next section.
 However, for fixed $n$ and $\mbox{rank}(\data)=d$, the complexity is exponential in $d$, which can not be improved unless $P=NP$ even for $m=2$ \cite{boob2018complexity}. 
 It is interesting to note that the proposed convex program trains ReLU neural networks \emph{optimally}, unlike local search heuristics such as backpropagation, which may converge to suboptimal solutions (see Section \ref{sec:numerical} for numerical evidence). To the best of our knowledge, our results provide the first polynomial-time algorithm to train non-trivial neural networks with global optimality guarantees.
 We also remark that further theoretical insight as well as faster numerical solvers can be developed due to the similarity to group Lasso \cite{yuan2006model} and related structured regularization methods. Theorem \ref{thm:mainconvex} implies that ReLU networks are equivalent to sparse mixtures of linear models, where sparsity is enforced by the group $\ell_1-\ell_2$ convex regularizer. More specifically, the non-convex neural network approach implicitly maps the data to the higher dimensional feature matrix $[D_1X, ..., D_PX]$, and consequently seeks a group sparse model.
\end{remark}
\begin{remark}%
We note that the convex program \eqref{eq:twolayerconvexprogram} can be approximated by sampling a set of diagonal matrices $D_1,...,D_{\tilde P}$. For example, one can generate $u\sim N(0,I_d)$, or from any distribution $\tilde P$ times, and let $D_i=\D(1[\data u_i\ge0])$, $\forall i \in [\tilde P]$ and solve the reduced convex problem, where remaining variables are set to zero. This is essentially a type of coordinate descent applied to \eqref{eq:twolayerconvexprogram}. In Section \ref{sec:numerical}, we show that this approximation in fact works extremely well, often better than backpropagation. In fact, backpropagation can be viewed as a heuristic method to solve the convex objective \eqref{eq:twolayerconvexprogram}. The global optima of this convex program \eqref{eq:twolayerconvexprogram} are among the fixed points of backpropagation, i.e., stationary points of \eqref{eq:2layer_regularized_cost}. Moreover, we can bound the suboptimality of any feasible solution, e.g., from backpropagation, in the non-convex cost \eqref{eq:2layer_regularized_cost} using the dual of \eqref{eq:twolayerconvexprogram}. 
\end{remark}
The proof of Theorem \ref{thm:mainconvex} can be found in Section \ref{sec:proofthm1}.
\section{Convolutional Neural Networks}
Here, we introduce extensions of our approach to convolutional neural networks (CNNs). Two-layer convolutional networks with $m$ hidden neurons (filters) of dimension $d$ and fully connected output layer weights can be described by patch matrices $X_k \in \real^{n\times d},\,k=1,...,K$. This formulation also includes image, or tensor inputs. For flattened activations, we have $f(\data_1,...,X_K)=\sum_{j=1}^m \sum_{k=1}^K  \phi(X_k\firstw_j)\secondw_{jk}$ as the network output. We first present a simpler case for vector regression, $f_k(\data_1,...,X_K)=\sum_{j=1}^m  \phi(X_k\firstw_{j})\secondw_{j}$ which is separable over the patch index $k$.

\subsection{Separable ReLU convolutional networks}
Consider the training problem
\begin{align*}
    \min_{\{\secondw_j,\firstw_j\}_{j=1}^m} \frac{1}{2} \sum_{k=1}^K \Big\| \sum_{j=1}^m (\data_k &\firstw_{j} )_+ \secondw_{j} - y_k \Big\|_2^2 \\
    &+\frac{\beta}{2} \sum_{j=1}^m (\|\firstw_{j}\|_2^2+\secondw_{j}^2)\,,
\end{align*}
where $y_k$'s are labels. We first note that this problem is separable over the patch indices $k$, therefore, do not exactly correspond to classical convolutional network architectures which are not separable.
The above problem can be reduced to the fully connected case \eqref{eq:2layer_regularized_cost} by defining $X^\prime=[X_1^T,...,X_K^T]^T$ and $y^\prime=[y_1^T,...,y_K^T]^T$. Therefore, the convex program \eqref{eq:twolayerconvexprogram} solves the above problem exactly in $O\Big(d^3r^3\big(\frac{n}{r}\big)^{3r}\Big)$ complexity, where $r$ is the number of variables in a single filter. Note that typical CNNs use $m$ filters of size $3\times 3$ (r=9) in the first layer \cite{he2016deep}. 
\subsection{Linear convolutional network training as a Semi-definite Program (SDP)}
We now start with the simple case of linear activations $\phi(t)=t$, where the training problem becomes
\begin{align}
\label{eq:CNN_linear}
    &\min_{\{\firstw_j,\secondw_j\}_{j=1}^{m}} \frac{1}{2}\Big\|\sum_{k=1}^K \sum_{j=1}^m  \data_k \firstw_j \secondw_{jk}-y \Big\|_2^2\\\nonumber
    &\hspace{4cm}+ \frac{\beta}{2} \sum_{j=1}^m \left(\|\firstw_j\|_2^2+\|\secondw_j\|_2^2 \right).
\end{align}
The corresponding dual problem is given by
\begin{align}
\label{eq:linearconvdual}
    &\max_{\dual} -\frac{1}{2}\|v-y\|_2^2+\frac{1}{2}\|y\|_2^2\,\,\mbox{s.t.} \max_{\|u\|_2\le 1}\, \sum_{k}\big( \dual^T \data_k u \big)^2 \le 1.
\end{align}
Similar arguments to those used in the proof of Theorem \ref{thm:mainconvex}, strong duality holds. Further, the maximizers of the inner problem are the maximal eigenvectors of $\sum_k \data_k^T \dual \dual^T \data_k$, which are optimal neurons (filters).
We can express \eqref{eq:linearconvdual} as the SDP
\begin{align}
    &\max_{\dual} -\frac{1}{2}\|v-y\|_2^2+\frac{1}{2}\|y\|_2^2 \nonumber\\
    &\mbox{s.t.  } \sigma_{\max}\left([\data_1^T \dual\, ...\, \data_K^T \dual ]\right)\le \beta.\label{eq:SDP}
\end{align}
The dual of the above SDP is a nuclear norm penalized convex optimization problem (see Appendix \ref{sec:sdp_appendix})
\begin{align}
\label{eq:CNN_linear_nuclear}
    &\min_{z_k\in\real^d ,\forall k  } \frac{1}{2}\Big\|\sum_{k=1}^K \data_k z_k-y \Big\|^2_2+ \beta\Big \|[z_1,\ldots,z_K] \Big \|_{*},
\end{align}
where $\Big\|[z_1,...,z_K] \Big \|_{*}=\|Z\|_*:=\sum_i \sigma_i(Z)$ is the $\ell_1$ norm of singular values, i.e., nuclear norm \cite{recht2010guaranteed}.  
\subsection{Linear circular convolutional networks}
Now, if we assume that the patches are padded with enough zeros and extracted with stride one, then the circular version of \eqref{eq:CNN_linear} can be written as
\begin{align}\label{eq:linear_cnn}
    \min_{\{\firstw_j,\secondw_j\}_{j=1}^m} \frac{1}{2} \Big\| \sum_{j=1}^m\data U_j \secondw_{j} -y \Big\|^2_2+\frac{\beta}{2}\sum_{j=1}^m \left( \|\firstw_j\|_2^2+\|\secondw_j\|_2^2 \right),
\end{align}
where $U_j \in \mathbb{R}^{d \times d}$ is a circulant matrix generated by a circular shift modulo $d$ using the elements $\firstw_j \in \mathbb{R}^h$. Then, the SDP \eqref{eq:SDP} reduces to (see Appendix \ref{sec:circular_linear_cnn_appendix})
\begin{align}\label{eq:linear_cnn_l1}
    \min_{z \in \mathbb{C}^d} \frac{1}{2}\Big \|\tilde{\data} z -y \Big\|^2_2+ \beta \|z\|_1,
\end{align}
where $\tilde{\data}=\data F$, and $F \in \mathbb{C}^{d \times d}$ is the Discrete Fourier Transform (DFT) matrix.

%

\section{Proof of the Main Result (Theorem \ref{thm:mainconvex})}
\label{sec:proofthm1}
We now prove the main result two-layer ReLU networks with squared loss\footnote{See Appendix \ref{sec:generalloss_appendix} for generic convex loss functions.}.
We start with the dual representation
\begin{align}
\label{eq:dual}
&\max_v \, -\frac{1}{2}\|v-y\|_2^2 + \frac{1}{2}\|y\|_2^2\\
&\mbox{s.t.}\, \max_{u:\,\|u\|_2\le 1} \, \vert v^T (\data u)_+ \vert \le \beta\,. \nonumber
\end{align}
Note that the constraint \eqref{eq:dual} can be represented as
\begin{align*}
\big\{v: \max_{\|u\|_2\le 1}   v^T (\data u)_+  \le \beta \big\}  \cap \big\{v:\max_{\|u\|_2\le 1}   -v^T (\data u)_+  \le \beta  \big\}.
\end{align*}

We now focus on a single-sided dual constraint
\begin{align}
\max_{u:\,\|u\|_2\le 1} \, v^T (\data u)_+ \le \beta, \label{eq:dualconst}
\end{align}
by considering hyperplane arrangements and a convex duality argument over each partition.
We first partition $\real^d$ into the following subsets
\begin{align*}
P_S := \{u\,:\, \datavec_i^T u &\ge 0, \forall i \in S,\,\,
\datavec_j^T u  \le 0, \forall j \in S^c \}.
\end{align*}
Let $\mathcal{H}_\data$ be the set of all hyperplane arrangement patterns for the matrix $\data$, defined as the following set
\begin{align*}
\mathcal{H}_\data = \bigcup \big \{ \{\sign(\data u)\} \,:\, u \in \real^d \big \}\,.
\end{align*}
It is obvious that the set $\mathcal{H}$ is bounded, i.e., $\exists N_H \in \mathbb{N} <\infty$ such that $\vert \mathcal{H} \vert \le N_H$. We next define an alternative representation of the sign patterns in $\mathcal{H}_\data$, which is the collection of sets that correspond to positive signs for each element in $\mathcal{H}$. More precisely, let
\begin{align*}
\mathcal{S}_X := \big\{ \{ \cup_{h_i=1} \{i\}  \} \,:\, h \in \mathcal{H}_\data \big\}.
\end{align*}
%
%

%
%
%
%

We now express the maximization in the dual constraint in \eqref{eq:dualconst} over all possible hyperplane arrangement patterns as
\begin{align*}
&\max_{u:\,\|u\|_2\le 1} \, v^T (\data u)_+ \\
&= \max_{u:\,\|u\|_2\le 1} v^T \D(\data u\ge 0) \data u\\
& = ~~\max_{\substack{S \subseteq [n]\\ S \in \mathcal{S}_\data }}\, \max_{\substack{u:\,\|u\|_2\le 1 \\ \datavec_i^T u \ge 0 \,\, \forall i \in S \\ \, \, \, \datavec_j^T u \le 0 \,\, \forall j \in S^c}} v^T\D(\data u\ge 0) \data u\\
& = ~~\max_{\substack{S \subseteq [n]\\ S \in \mathcal{S}_\data }} \max_{\substack{u:\,\|u\|_2\le 1 \\ u \in P_S}} v^T\D(\data u\ge 0) \data u
\end{align*}
Let us define the diagonal matrix $D(S)\in \real^{n\times n}$ which is a function of the subset $S \subseteq [n]$.
\begin{align*}
D(S)_{ii} := \begin{cases} 1 & \mbox{ if } i \in S\\
0 & \mbox{ otherwise.}  \end{cases}
\end{align*}
Note that $D(S^c) = I_n-D(S)$, since $S^c$ is the complement of the set $S$.
With this notation, we can represent $P_S$ as
\begin{align*}
P_S = \{ u\,:\, D(S)\data u \ge 0,\,\, (I_n-D(S))u\le 0 \}\,,
\end{align*}
%
and the maximization in the dual constraint as
\begin{align*}
\max_{u:\,\|u\|_2\le 1} \, v^T (\data u)_+ 
= ~~\max_{\substack{S \subseteq [n]\\ S \in \mathcal{S}_\data }} \max_{\substack{~u:\,\|u\|_2\le 1 \\ u \in P_S}} v^TD(S) \data u\,.
\end{align*}
Enumerating all hyperplane arrangements $\mathcal{H}_\data$, or equivalently $\mathcal{S}_\data$, we index them in an arbitrary order via $i \in [\vert S_\data \vert]$. We denote $M=\vert S_\data \vert$. Hence, $S_1,...,S_M \in \mathcal{S}_\data$ is the list of all $M$ elements of $\mathcal{S}_\data$. 
Next we use the strong duality result from Lemma \ref{lem:finitestrongdualconst} for the inner maximization problem.
The dual constraint \eqref{eq:dualconst} can be represented as 
\begin{align*}
\eqref{eq:dualconst} &\iff \forall i\in [M],\, \mbox{ it holds that }\\
  &\qquad \, \min_{\substack{\alpha, \beta \in \real^{n} \\ \alpha, \beta \ge 0}}
\| \data^T D(S_i) \big(v + \alpha + \beta \big)-\data^T \beta \|_2 \le \beta\\
& \iff \forall i\in [M],\, \exists \alpha_i,\beta_i \in {\real^n} \mbox{ s.t.} \\
&\qquad \quad \alpha_i,\beta_i \ge 0\\
&\qquad \quad \| \data^T D(S_i) \big(v + \alpha_i + \beta_i \big)-\data^T \beta_i \|_2 \le \beta\,.
\end{align*}
Therefore, recalling the two-sided constraint in \eqref{eq:dual}, we can represent the dual optimization problem in \eqref{eq:dual} as a finite dimensional convex optimization problem with variables $v \in \real^n, \alpha_i,\beta_i,\alpha_i^\prime,\beta_i^\prime \in  {\real^n}, \forall i \in [M]$, and $2M$ second order cone constraints as follows
\begin{align*}
&\max_{\substack{v\in \real^n \\ \alpha_i,\beta_i \in {\real^n}\\\alpha_i, \beta_i\ge 0, \, \forall i \in [M] \\ \alpha_i^\prime,\beta_i^\prime \in {\real^n}\\\alpha_i^\prime, \beta_i^\prime\ge 0, \, \forall i \in [M] }}\, -\frac{1}{2}\|v-y\|_2^2 + \frac{1}{2}\|y\|_2^2 
\\
&\mbox{s.t.}\, \| \data^T D(S_1) \big(v + \alpha_1 + \beta_1 \big)-\data^T \beta_1 \|_2 \le \beta \nonumber\\ 
& \qquad \vdots \nonumber\\
& \quad\,\, \| \data^T D(S_M) \big(v + \alpha_M + \beta_M \big)-\data^T \beta_M \|_2 \le \beta \nonumber\\
& \quad\,\, \| \data^T D(S_1) \big(-v + \alpha_1^\prime + \beta_1^\prime \big)-\data^T \beta_1^\prime \|_2 \le \beta \nonumber\\
& \qquad \vdots \nonumber\\
& \quad\,\, \| \data^T D(S_M) \big(-v + \alpha_M^\prime + \beta_M^\prime \big)-\data^T \beta_M^\prime \|_2 \le \beta. \nonumber\\
\end{align*}
The above problem can be represented as a standard finite dimensional second order cone program. Note that the particular choice of parameters $v=\alpha_i=\beta_i=\alpha_i^\prime=\beta_i^\prime=0$, $\forall i \in [M]$, are strictly feasible in the above constraints as long as $\beta>0$. Therefore Slater's condition and consequently strong duality holds \cite{Boyd02}. The dual problem \eqref{eq:dual} can be written as

\allowdisplaybreaks
\begin{align*}
&\min_{\substack{\lambda, \lambda^\prime \in \real^M\\ \lambda,\lambda^\prime\ge 0}} \, \max_{\substack{v\in \real^n \\ \alpha_i,\beta_i \in {\real^n}\\\alpha_i, \beta_i\ge 0, \, \forall i \in [M]\\ \alpha_i^\prime,\beta_i^\prime \in {\real^n}\\\alpha_i^\prime, \beta_i^\prime\ge 0, \, \forall i \in [M] }}\, -\frac{1}{2}\|v-y\|_2^2 + \frac{1}{2}\|y\|_2^2 \\
 &+ \sum_{i=1}^M \lambda_i \big(\beta - \| \data^T D(S_i) \big(v + \alpha_i + \beta_i \big)-\data^T \beta_i \|_2 \big)\\
  &+ \sum_{i=1}^M \lambda_i^\prime \big(\beta - \| \data^T D(S_i) \big(-v + \alpha_i^\prime + \beta_i^\prime \big)-\data^T \beta_i^\prime \|_2 \big). 
\end{align*}
Next we introduce variables $r_1,\ldots,r_M, r_1^\prime,\ldots,r_M^\prime \in\real^d$ and represent the dual problem \eqref{eq:dual} as
\allowdisplaybreaks
\begin{align*}
&\min_{\substack{\lambda,\lambda^\prime \in \real^M\\ \lambda,\lambda^\prime\ge 0}} \, \max_{\substack{v\in \real^n \\ \alpha_i,\beta_i \in {\real^n}\\\alpha_i, \beta_i\ge 0, \,\forall i  \\ \alpha_i^\prime,\beta_i^\prime \in {\real^n}\\\alpha_i^\prime, \beta_i^\prime\ge 0, \, \forall i }}\, \min_{\substack{r_i \in \real^d,\, \|r_i\|_2\le 1\\ r_i^\prime \in \real^d,\, \|r_i^\prime\|_2\le 1\\ \forall i \in [M]} } -\frac{1}{2}\|v-y\|_2^2 + \frac{1}{2}\|y\|_2^2 \\
 &+ \sum_{i=1}^M \lambda_i \big(\beta + r_i^T \data^T D(S_i) \big(v + \alpha_i + \beta_i \big)-r_i^T\data^T \beta_i  \big)\\
 &\hspace{-0.3cm}+ \sum_{i=1}^M \lambda_i^\prime \big(\beta + {r_i^\prime}^T \data^T D(S_i) \big(-v + \alpha_i^\prime + \beta_i^\prime \big)-{r_i^\prime}^T\data^T \beta_i^\prime  \big)\,.
\end{align*}
%
We note that the objective is concave in $v,\alpha_i,\beta_i$ and convex in $r_i, r_i^\prime$, $\forall i \in [M]$. Moreover the constraint sets $\|r_i\|_2\le 1,\, \|r_i^\prime\|_2\le 1, \,\forall i$ are convex and compact. Invoking Sion's minimax theorem \cite{sion_minimax} for the inner $\max \min$ problem, we may express the strong dual of the problem \eqref{eq:dual} as
\begin{align*}
&\min_{\substack{\lambda,\lambda^\prime \in \real^M\\ \lambda,\lambda^\prime\ge 0}} \,\min_{\substack{r_i \in \real^d,\, \|r_i\|_2\le 1\\ r_i^\prime \in \real^d,\, \|r_i^\prime\|_2\le 1} } \max_{\substack{v\in \real^n \\  \alpha_i,\beta_i \in {\real^n}\\\alpha_i, \beta_i\ge 0, \, \forall i  \\ \alpha_i^\prime,\beta_i^\prime \in {\real^n}\\\alpha_i^\prime, \beta_i^\prime\ge 0, \, \forall i }}\,  -\frac{1}{2}\|v-y\|_2^2 + \frac{1}{2}\|y\|_2^2 \\
 &+ \sum_{i=1}^M\lambda_i \big(\beta + r_i^T \data^T D(S_i) \big(v + \alpha_i + \beta_i \big)-r_i^T\data^T \beta_i  \big)\\
  &\hspace{-0.3cm}+ \sum_{i=1}^M \lambda_i^\prime \big(\beta + {r_i^\prime}^T \data^T D(S_i) \big(-v + \alpha_i^\prime + \beta_i^\prime \big)-{r_i^\prime}^T\data^T \beta_i^\prime  \big)\,.
\end{align*}

\begin{figure*}[ht]
\centering
\captionsetup[subfigure]{oneside,margin={1cm,0cm}}
	\begin{subfigure}[t]{0.32\textwidth}
	\centering
	\includegraphics[width=1.11\textwidth, height=0.8\textwidth]{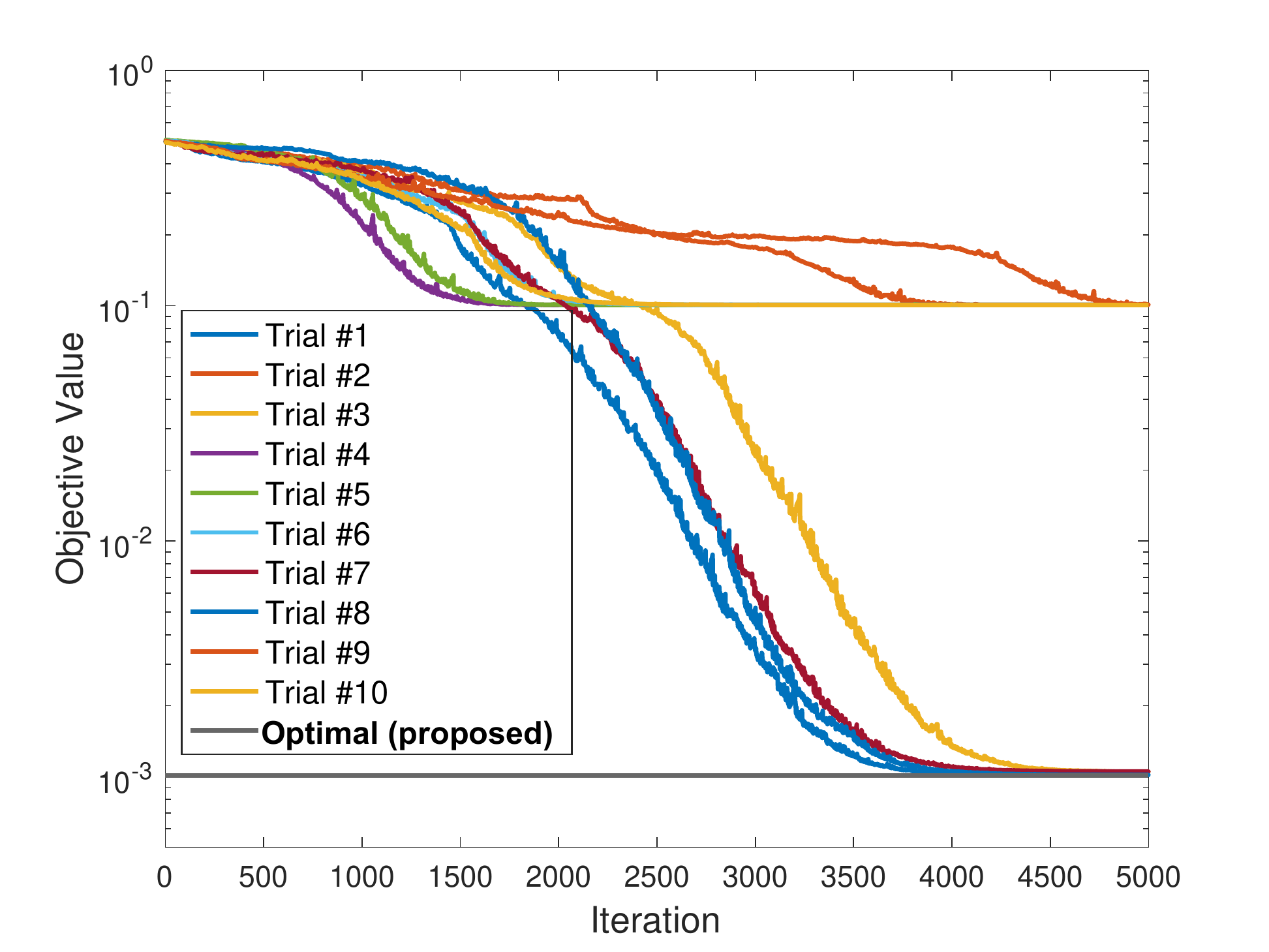}
	\caption{$m=8$\centering} \label{fig:sgd_m=8}
\end{subfigure} \hspace*{\fill}
	\begin{subfigure}[t]{0.32\textwidth}
	\centering
	\includegraphics[width=1.11\textwidth, height=0.8\textwidth]{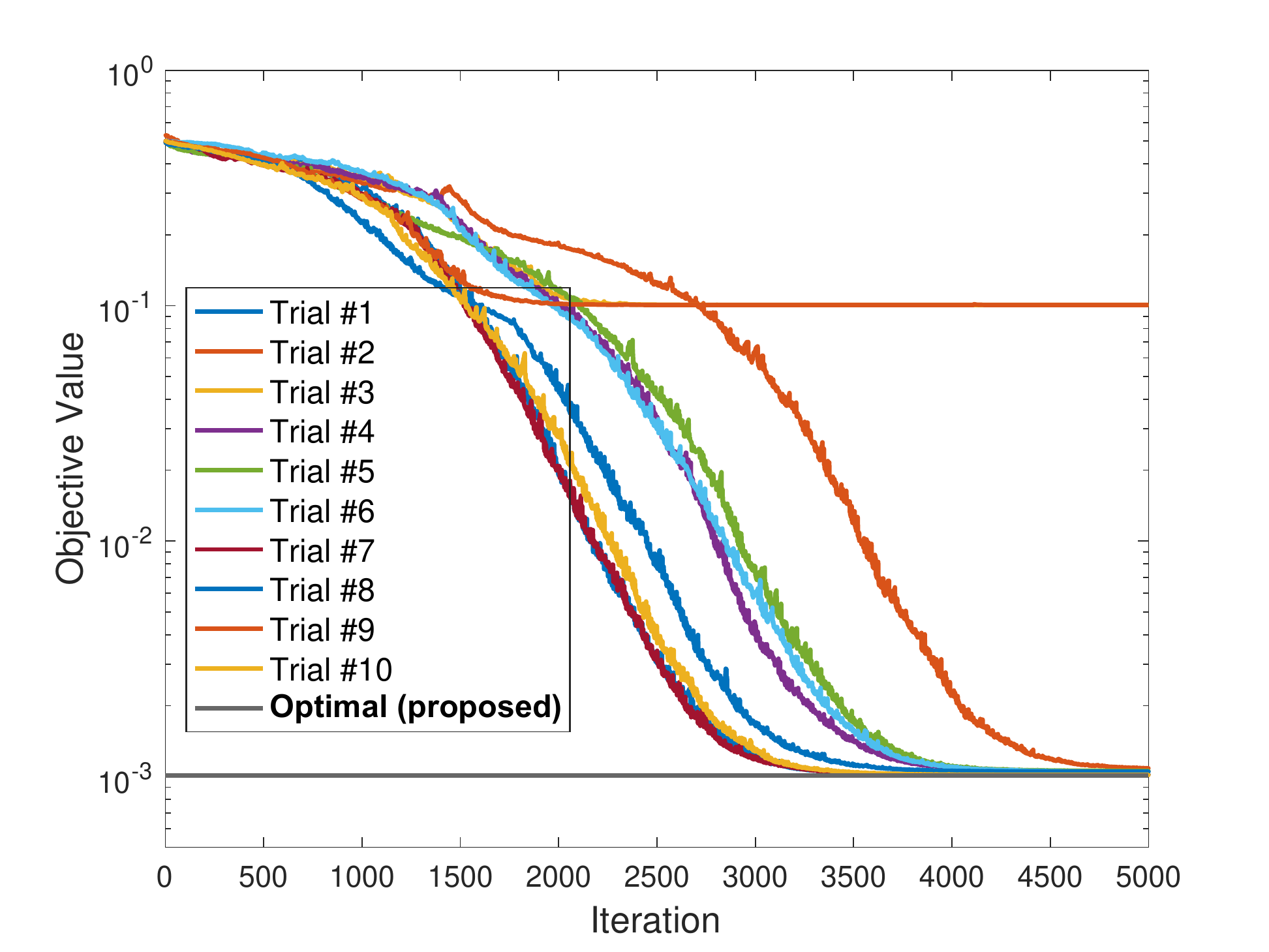}
	\caption{$m=15$\centering} \label{fig:sgd_m=15}
\end{subfigure} \hspace*{\fill}
	\begin{subfigure}[t]{0.32\textwidth}
	\centering
	\includegraphics[width=1.11\textwidth, height=0.8\textwidth]{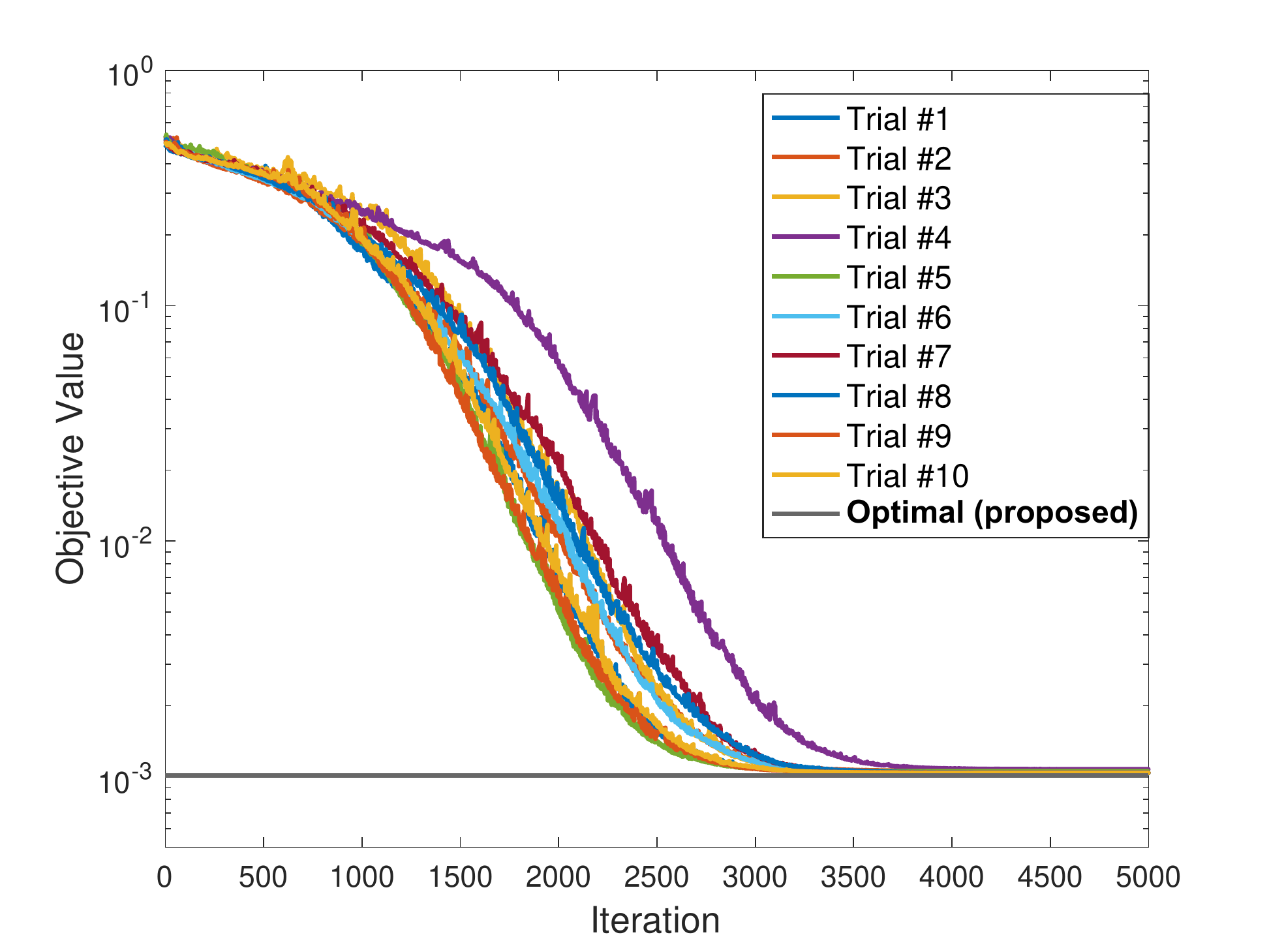}
	\caption{$m=50$\centering} \label{fig:sgd_m=50}
\end{subfigure} \hspace*{\fill}
\caption{Training cost of a two-layer ReLU network trained with SGD (10 initialization trials) on a one dimensional dataset ($d=1$), where Optimal denotes proposed convex programming approach in \eqref{eq:twolayerconvexprogram}. SGD can be stuck at local minima for small $m$, while the proposed approach is optimal as guaranteed by Theorem \ref{thm:mainconvex}. }\label{fig:sgd_1d}
\vskip -0.1in
\end{figure*}

\begin{figure*}[ht]
\centering
\captionsetup[subfigure]{oneside,margin={1cm,0cm}}
	\begin{subfigure}[t]{0.45\textwidth}
	\centering
	\includegraphics[width=1.0\textwidth, height=0.8\textwidth]{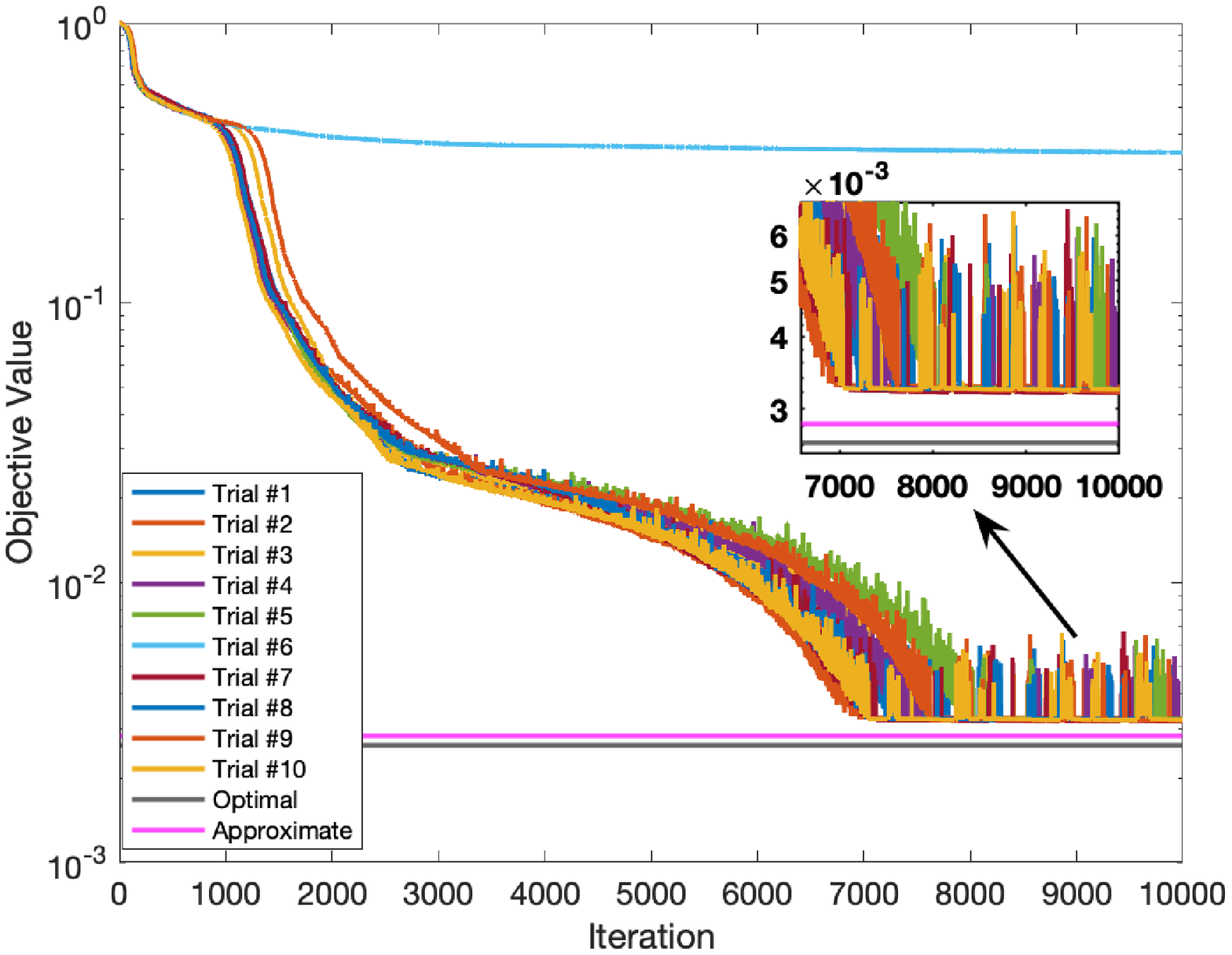}
	\caption{Independent realizations with $m=50$\centering} \label{fig:trials_minibatch_2d}
\end{subfigure} \hspace*{\fill}
	\begin{subfigure}[t]{0.45\textwidth}
	\centering
	\includegraphics[width=1.11\textwidth, height=0.8\textwidth]{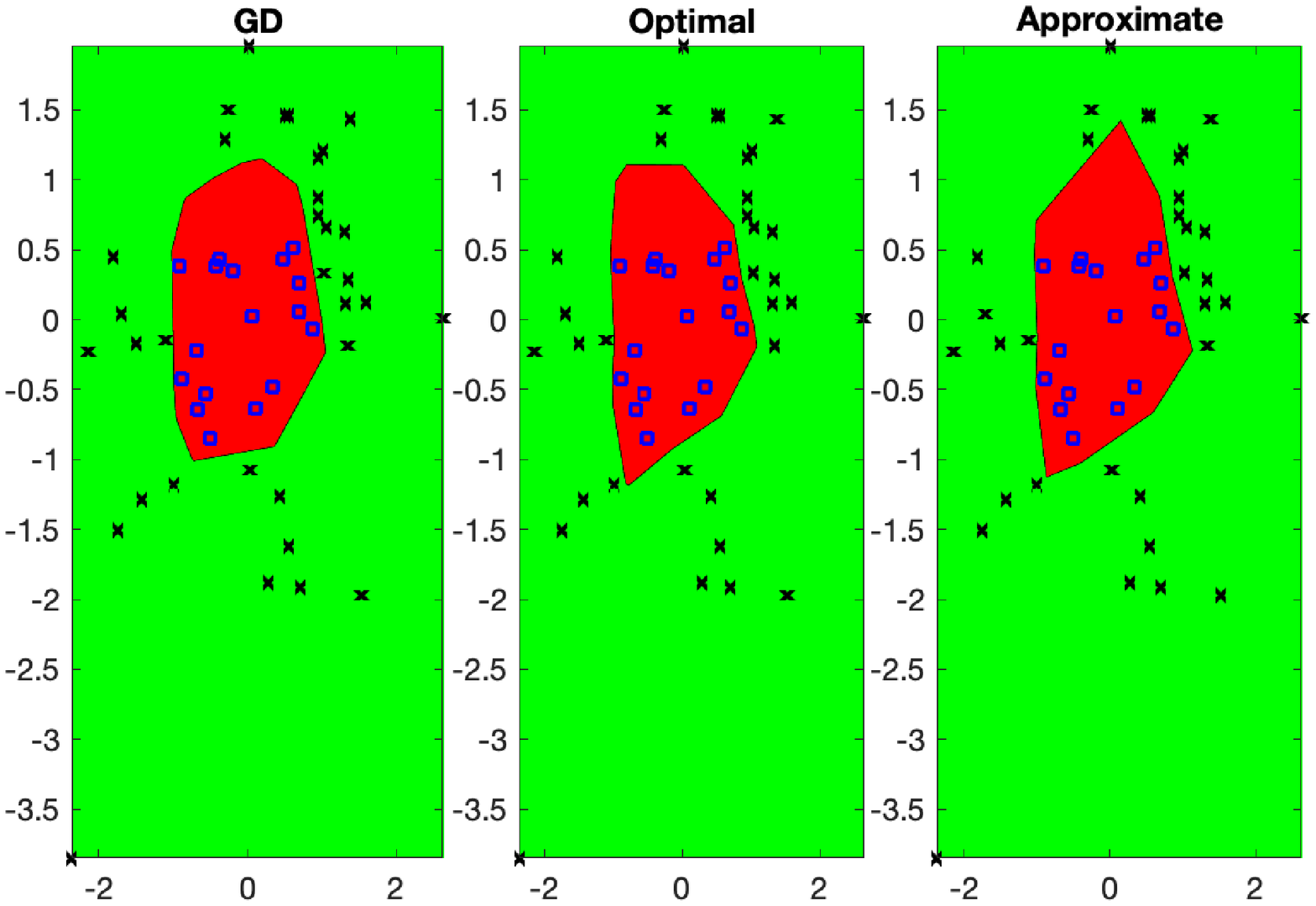}
	\caption{Decision boundaries\centering} \label{fig:hinge_minibatch_2d}
\end{subfigure} \hspace*{\fill}
\caption{Training cost of a two-layer ReLU network trained with SGD (10 initialization trials) on a two-dimensional dataset, where Optimal and Approximate denote the objective value obtained by the proposed convex program in \eqref{eq:twolayerconvexprogram} and its approximation by sampling variables, respectively. Learned decision boundaries are also depicted. }\label{fig:minibatch_2d}
\vskip -0.1in
\end{figure*}

\begin{figure*}[ht]
\centering
\captionsetup[subfigure]{oneside,margin={1cm,0cm}}
	\begin{subfigure}[t]{0.45\textwidth}
	\centering
	\includegraphics[width=1.11\textwidth, height=0.8\textwidth]{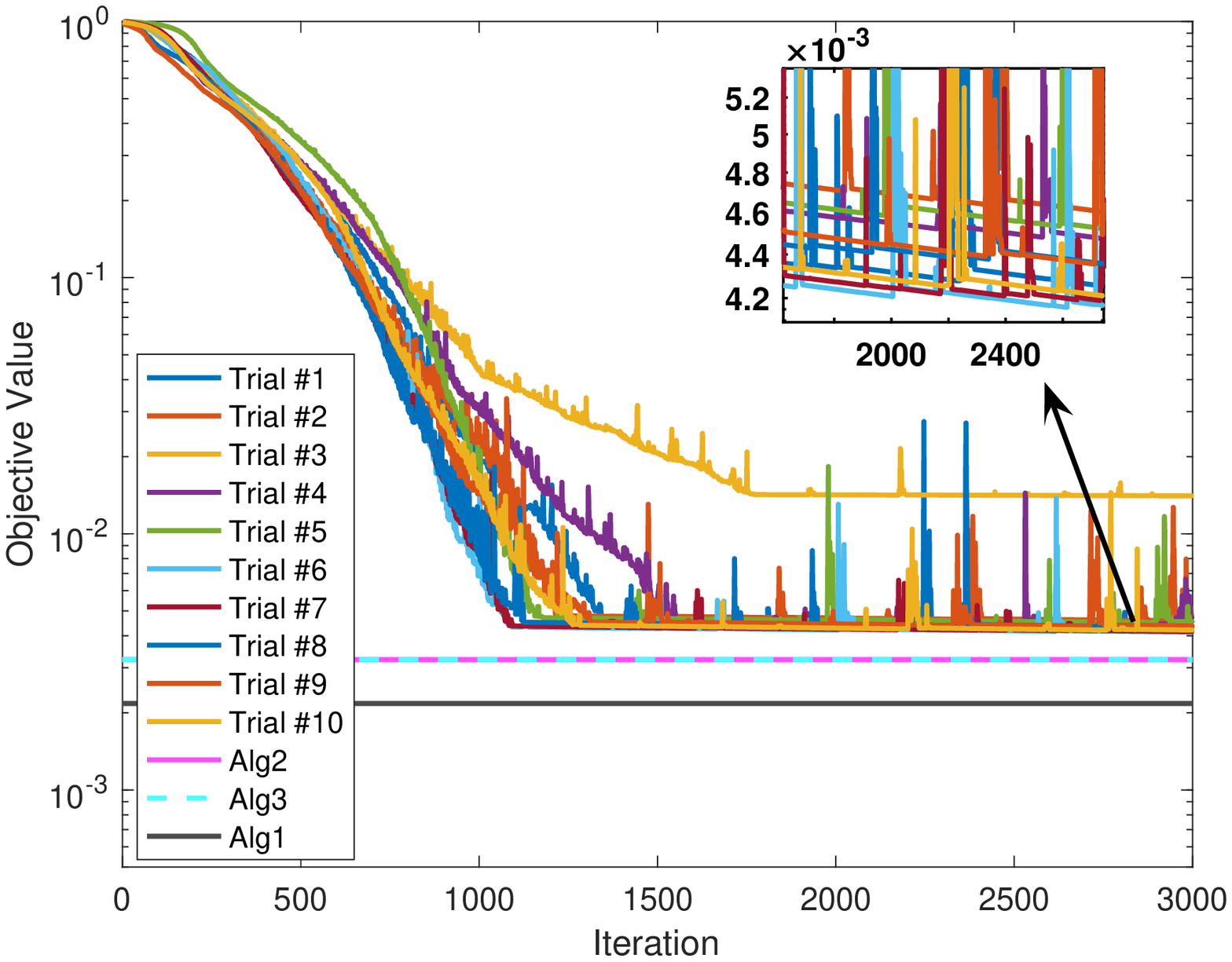}
	\caption{Objective value\centering} \label{fig:relu_cifar_obj}
\end{subfigure} \hspace*{\fill}
	\begin{subfigure}[t]{0.45\textwidth}
	\centering
	\includegraphics[width=1.11\textwidth, height=0.8\textwidth]{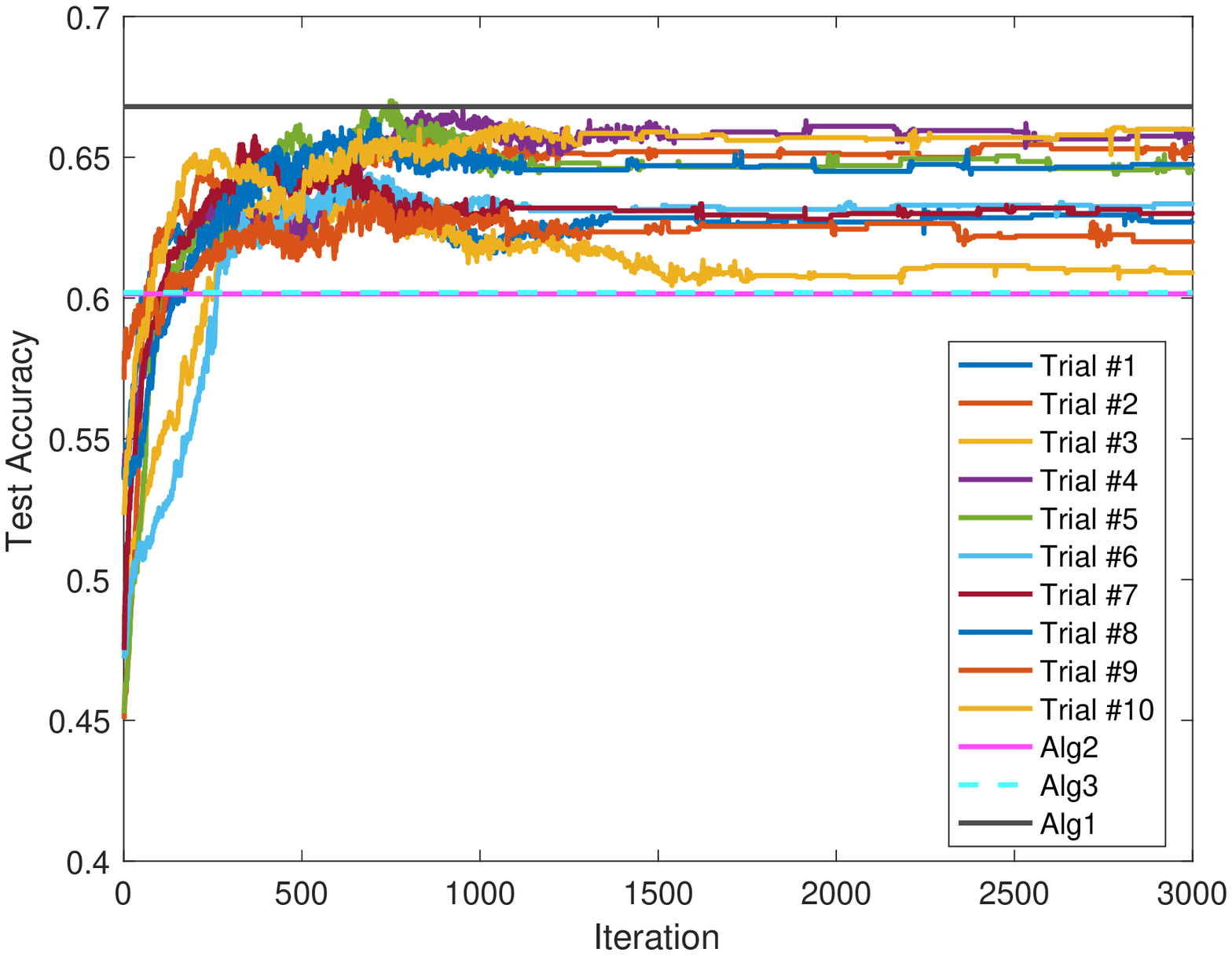}
	\caption{Test accuracy\centering} \label{fig:relu_cifar_acc}
\end{subfigure} \hspace*{\fill}
\caption{Training cost of a two-layer ReLU network trained with SGD (10 initialization trials) on a subset of CIFAR-10 and the convex program \eqref{eq:twolayerconvexprogram} denoted as Alg1. Alg2 and Alg3, which are approximations of the convex program.}\label{fig:relu_cifar}
\vskip -0.15in
\end{figure*}
\begin{figure*}[ht]
\centering
\captionsetup[subfigure]{oneside,margin={1cm,0cm}}
	\begin{subfigure}[t]{0.45\textwidth}
	\centering
	\includegraphics[width=1\textwidth, height=0.8\textwidth]{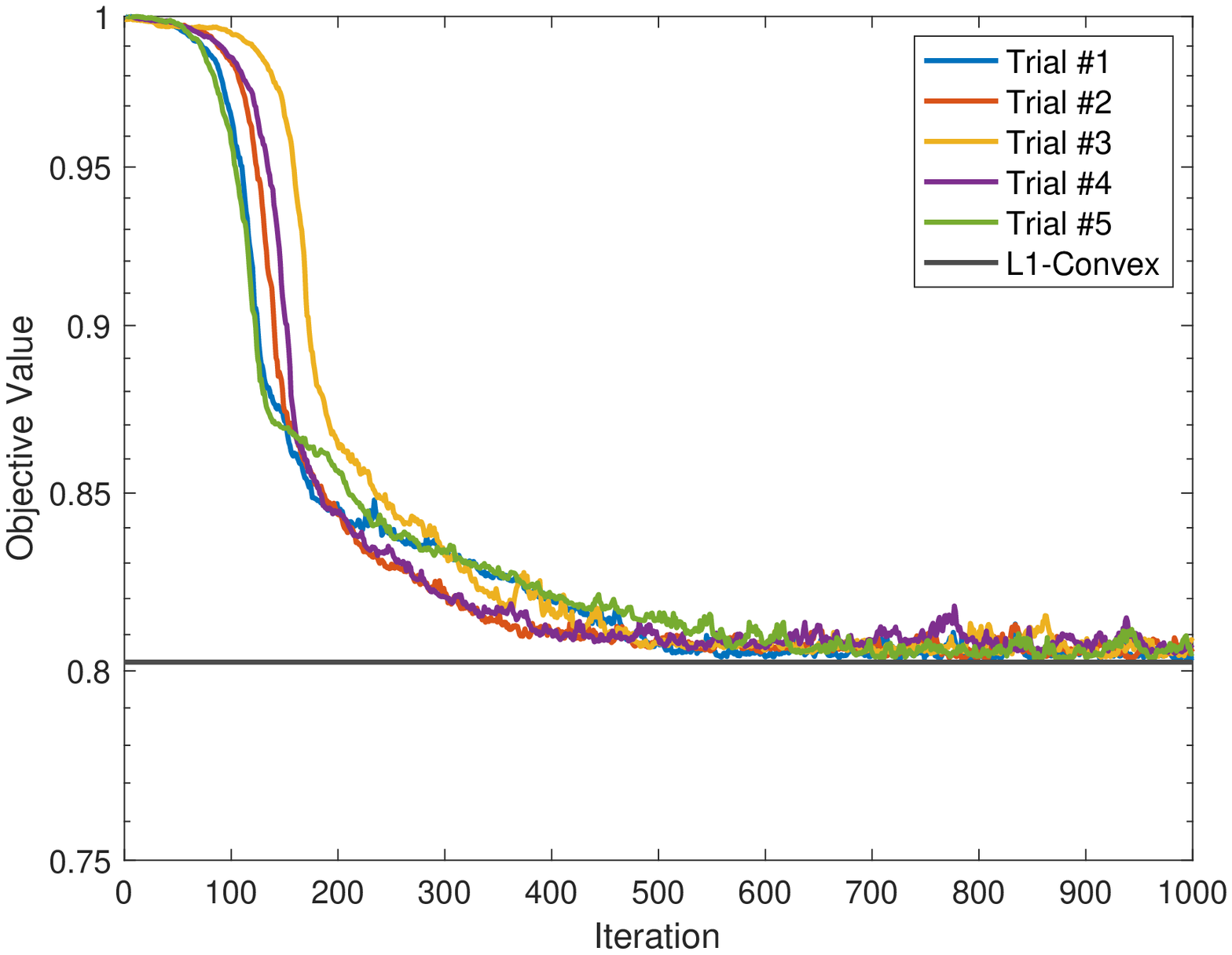}
	\caption{Objective value\centering} \label{fig:linear_cnn_obj}
\end{subfigure} \hspace*{\fill}
	\begin{subfigure}[t]{0.45\textwidth}
	\centering
	\includegraphics[width=1\textwidth, height=0.8\textwidth]{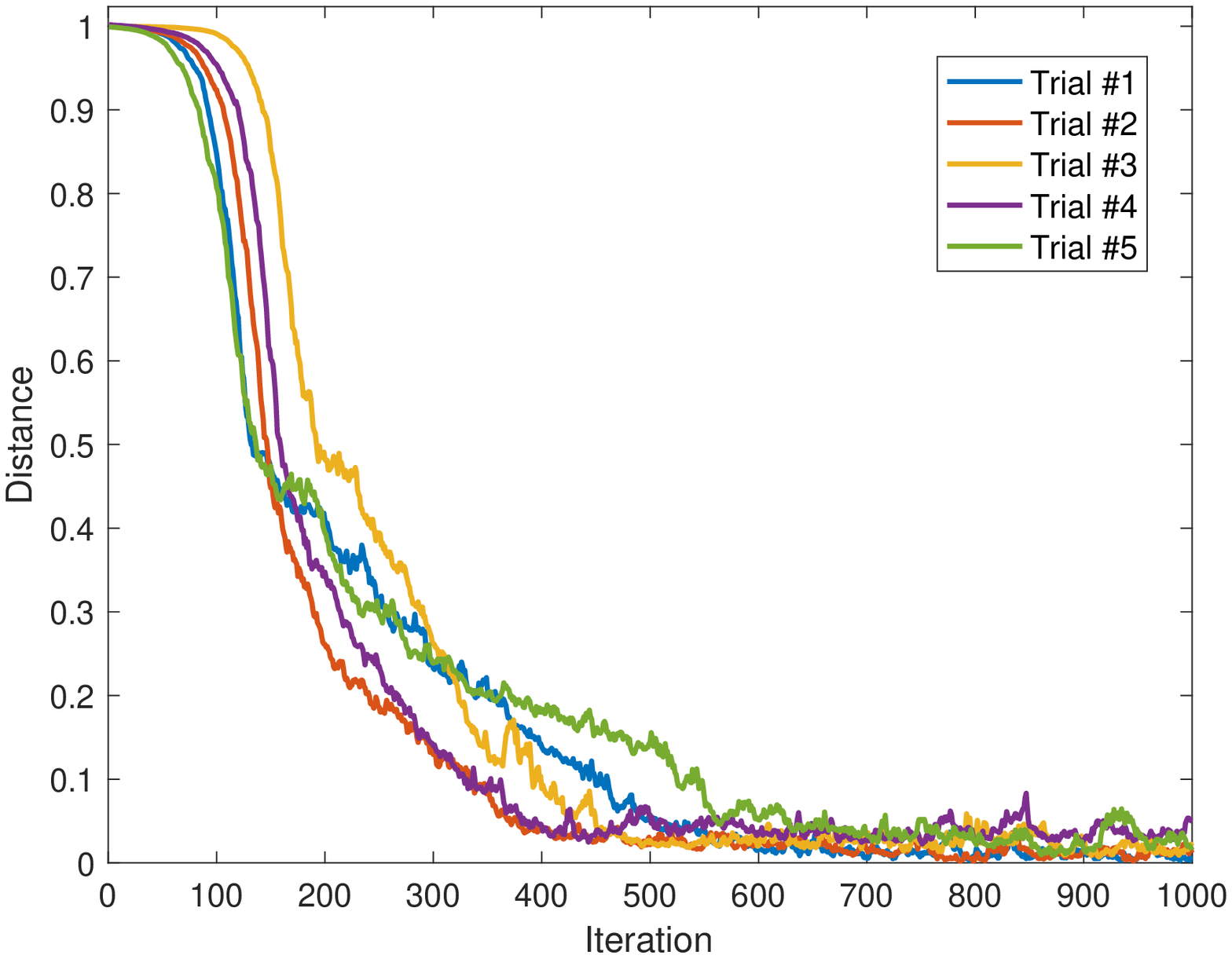}
	\caption{Distance to the solution of the convex program \centering} \label{fig:linear_cnn_distance}
\end{subfigure} \hspace*{\fill}
\caption{Training accuracy of a two-layer linear CNN trained with SGD (5 initialization trials) on a subset of CIFAR-10, where L1-Convex denotes the proposed convex program in \eqref{eq:linear_cnn_l1}. Filters found via SGD converge to the solution of \eqref{eq:linear_cnn_l1}. }\label{fig:linear_cnn}
\vskip -0.2in
\end{figure*}

%
Computing the maximum with respect to $v,\alpha_i,\beta_i,\alpha_i^\prime,\beta_i^\prime$, $\forall i \in [M]$, analytically we obtain the strong dual to \eqref{eq:dual} as
\begin{align*}
&\min_{\substack{\lambda,\lambda^\prime \in \real^M\\ \lambda,\lambda^\prime\ge 0}} \,\min_{\substack{r_i \in \real^d,\, \|r_i\|_2\le 1\\r_i^\prime \in \real^d,\, \|r_i^\prime\|_2\le 1 \\  (2D(S_i)-I_n)Xr_i\ge 0\\  (2D(S_i)-I_n)Xr_i^\prime\ge 0}}
  {\frac{1}{2}}\Big\|\sum_{i=1}^M \lambda_i D(S_i) \data r_i^\prime\\
 &-\lambda_i^\prime D(S_i) \data r_i - y \Big\|_2^2
 + \beta \sum_{i=1}^M (\lambda_i+\lambda_i^\prime).
\end{align*}
Now we apply a change of variables and define $w_i = \lambda_i r_i$ and $w_i^\prime=\lambda_i^\prime r_i^\prime$, $\forall i \in [M]$. Note that we can take $r_i=0$ when $\lambda_i=0$ without changing the optimal value. We obtain
\begin{align*}
&\hspace{-0.5cm}\min_{\substack{w_i,w_i^\prime \in P_{S_i}\\ \|w_i\|_2\le \lambda_i\\
 \|w_i^\prime\|_2\le \lambda_i^\prime \\ \lambda,\lambda^\prime\ge 0}}
  \frac{1}{2}\Big\|\sum_{i=1}^M D(S_i) \data (w_i^\prime-w_i) - y \Big\|_2^2 
+ \beta \sum_{i=1}^M (\lambda_i+\lambda_i^\prime).
\end{align*}
The variables $\lambda_i$, $\lambda_i^\prime$, $\forall i \in [M]$ can be eliminated since $\lambda_i=\|w_i\|_2$ and $\lambda_i^\prime=\|w_i^\prime\|_2$ are feasible and optimal. Plugging in these expressions, we get
\begin{align*}
&\min_{\substack{w_i,w_i^\prime\in P_{S_i}}}
  {\frac{1}{2}}\Big\|\sum_{i=1}^M D(S_i) \data (w_i^\prime-w_i) - y \Big\|_2^2 
\\&+ \beta \sum_{i=1}^M (\|w_i\|_2+\|w_i^\prime\|_2)\,,
\end{align*}
which is identical to \eqref{eq:twolayerconvexprogram}, and proves that the objective values are equal. 
Constructing $\{u^*_j,\alpha^*_j\}_{j=1}^{m^*}$ as stated in the theorem, and plugging in \eqref{eq:2layer_regularized_cost}, we obtain the value
%
\begin{align*}
    p^*\le&\frac{1}{2} \Big \| \sum_{j=1}^{m^*} (Xu_j^*)_+\alpha_j^*- y \Big\|_2^2 \\
    &+ \frac{\beta}{2} \sum_{i=1, v^*_i\neq 0}^{P} \left( \Big\|\frac{v^*_i}{\sqrt{\|v^*_i\|_2}}\Big\|_2^2 + \Big\|\sqrt{\|v^*_i\|_2} \Big\|_2^2 \right) \\
    &+ \frac{\beta}{2} \sum_{i=1, w^*_i\neq 0}^{P} \left(\Big\|\frac{w^*_i}{\sqrt{\|w^*_i\|_2}}\Big\|_2^2 + \Big\|\sqrt{\|w^*_i\|_2} \Big\|_2^2\right)
,\end{align*}
which is identical to the objective value of the convex program \eqref{eq:twolayerconvexprogram}. Since the value of the convex program is equal to the value of it's dual $d^*$ in \eqref{eq:dual}, and $p^*\ge d^*$, we conclude that $p^*=d^*$, which is equal to the value of the convex program \eqref{eq:twolayerconvexprogram} achieved by the prescribed parameters.
\qed


\section{Numerical Experiments} \label{sec:numerical}
In this section, we present small scale numerical experiments to verify our results in the previous sections\footnote{Additional experiments can be found in the Appendix \ref{sec:experiments_appendix}}. We first consider a one-dimensional dataset with $n=5$, i.e., $\data=[-2 \; -1 \; 0 \; 1 \; 2]^T$ and $y=[1 \;-1 \; 1 \; 1 \; -1]^T$, where we include the bias term by simply concatenating a column of ones to the data matrix $X$. We then fit these data points using a two-layer ReLU network trained with SGD and the proposed convex program, where we use squared loss as a performance metric. In Figure \ref{fig:sgd_1d}, we plot the value of the regularized objective function with respect to the iteration index. Here, we plot $10$ independent realizations for SGD and denote the convex program in \eqref{eq:twolayerconvexprogram} as ``Optimal''. Additionally, we repeat the same experiment for different number of neurons, particularly, $m=8,15$, and $50$. As demonstrated in the figure, when the number of neurons is small, SGD is stuck at local minima. As we increase $m$, the number of trials that achieve the optimal performance gradually increases as well, which is also consistent with the interpretations in \cite{ergen2019shallow}. We also note that Optimal achieves the smallest objective value as claimed in the previous sections. We then compare the performances on two-dimensional datasets with $n=50$, $m=50$ and $y \in \{+1,-1\}^n$, where we use SGD with the batch size $25$ and hinge loss as a performance metric. In these experiments, we also consider an approximate convex program, i.e., denoted as ``Approximate'' for which we use only a random subset of the diagonal matrices $D_1,...D_P$ of size $m$. As illustrated in Figure \ref{fig:minibatch_2d}, most of the SGD realizations converge to a slightly higher objective than Optimal. Interestingly, we also observe that even Approximate can outperform SGD in this case. In the same figure, we also provide the decision boundaries obtained by each method. 

We also evaluate the performance of the algorithms on a small subset of CIFAR-10 for binary classification \cite{cifar10}. Particularly, in each experiment, we first select two classes and then randomly under-sample to create a subset of the original dataset. For these experiments, we use hinge loss and SGD. In the first experiment, we train a two-layer ReLU network on the subset of CIFAR-10, where we include three different versions denoted as ``Alg1'', ``Alg2'', and ``Alg3'', respectively. For Alg1, we use a random subset of the diagonal matrices $D_1,...,D_p$ which match the sign patterns of the optimized (by SGD) network along with a randomly selected subset of possible sign patterns. Similarly, for Alg2, we use the sign patterns that match the initialized network. For Alg3, we perform a heuristic adaptive sampling for the diagonal matrices: we first examine the values of $\data \firstw$ for each neuron using the initial weights and flip the sign pattern corresponding to small values and use it along with the original sign pattern. In Figure \ref{fig:relu_cifar}, we plot both the objective value and the corresponding test accuracy for $10$ independent realizations with $n=106$, $d=100$, $m=12$, and batch size $25$. We observe that Alg1 achieves the lowest objective value and highest test accuracy.
Finally, we train a two-layer linear CNN architecture on a subset of CIFAR-10, where we denote the proposed convex program in \eqref{eq:linear_cnn_l1} as ``L1-Convex''. In Figure \ref{fig:linear_cnn}, we plot both the objective value and the Euclidean distance between the filters found by GD and L1-Convex for $5$ independent realizations with $n=387$, $m=30$, $h=10$, and batch size $60$. In this experiment, all the realizations converge to the objective value obtained by L1-Convex and find almost the same filters.

\section{Concluding Remarks}
We introduced a convex duality theory for non-convex neural network objectives and developed an exact representation via a convex program with polynomial many variables and constraints. Our results provide an equivalent characterization of neural network models in terms of convex regularization in a higher dimensional space where the data matrix is partitioned over all possible hyperplane arrangements. ReLU neural networks can be precisely represented as convex regularizers, where piecewise linear models are fitted via an $\ell_1-\ell_2$ group norm regularizer. It is well known that two-layer networks have a quite rich representation power thanks to their universal approximation property. However, our results clearly show that the fitted model is parsimonious due to the $\ell_1-\ell_2$ group regularization, which facilitates better generalization. Thus, we believe that our characterization sheds light into the extraordinary success of ReLU networks. There are a multitude of open research directions. One can obtain a better understanding of neural networks and their generalization properties by leveraging convexity, and high dimensional regularization theory \cite{wainwright2019high}. In the light of our results, one can view backpropagation as a heuristic method to solve the convex program \eqref{eq:twolayerconvexprogram} and analyze the loss landscape, since the global minima are necessarily stationary points of the non-convex objective \eqref{eq:2layer_regularized_cost}, i.e., fixed points of the update rule. Interesting consequences in this direction are reported in \cite{lacotte2020local} after our work. Furthermore, one can extend our convex approach to various architectures, e.g., modern CNNs, recurrent networks, and autoencoders. Based on our methods, recently \cite{ergen2020cnn} considered convex programs for CNNs with various pooling strategies, and determined several other convex regularizers implied by the network architecture. Finally, to the best of our knowledge, our results provide the first algorithm to train non-trivial neural networks \emph{optimally}. On the other hand, the popular backpropagation method is a local search heuristic, which may not find the optimal neural network and may be dramatically inefficient as shown in the experiments. Efficient optimization algorithms that exactly or approximately solve the convex program can be developed for larger scale experiments, including proximal and stochastic gradient methods.

\section*{Acknowledgements}
This work was supported in part by the National Science Foundation under grant IIS-1838179 and Stanford SystemX Alliance.

\bibliographystyle{icml2020}
\bibliography{references}

%
%
\clearpage
\onecolumn
\appendix
\addcontentsline{toc}{section}{Appendix} 
\part{Appendix} 
\parttoc 
\section{Appendix}
\subsection{Additional numerical results}\label{sec:experiments_appendix}
\begin{figure*}[ht!]
\centering
\captionsetup[subfigure]{oneside,margin={1cm,0cm}}
	\begin{subfigure}[t]{0.45\textwidth}
	\centering
	\includegraphics[width=1.11\textwidth, height=0.8\textwidth]{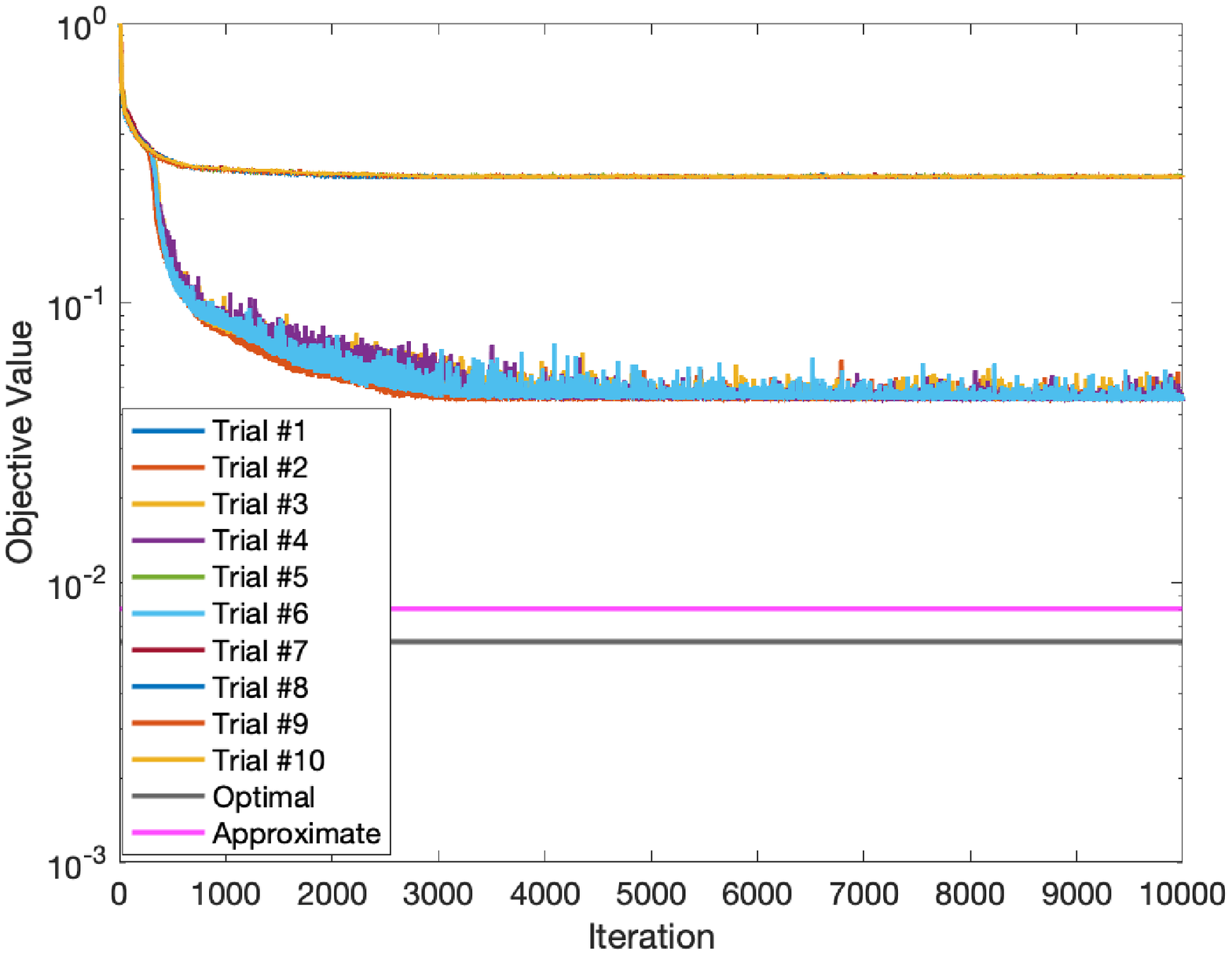}
	\caption{Independent SGD initialization trials with $m=50$\centering} \label{fig:trials_minibatch_2d_anomaly}
\end{subfigure} \hspace*{\fill}
	\begin{subfigure}[t]{0.45\textwidth}
	\centering
	\includegraphics[width=1.11\textwidth, height=0.8\textwidth]{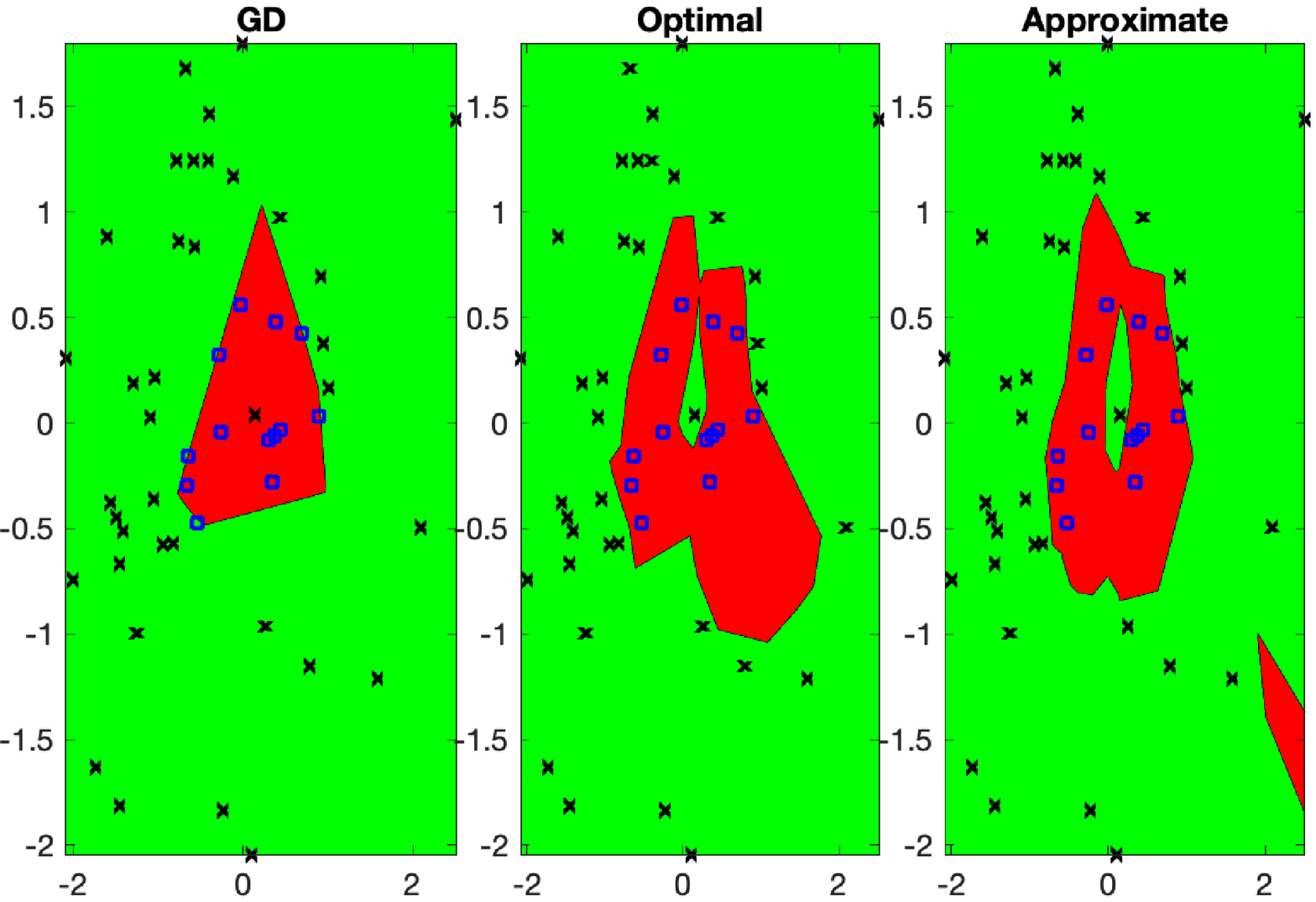}
	\caption{Decision boundaries\centering} \label{fig:hinge_minibatch_2d_anomaly}
\end{subfigure} \hspace*{\fill}
\caption{Training of a two-layer ReLU network with SGD (10 initialization trials) and proposed convex programs on a two-dimensional dataset. Optimal and Approximate denote the objective value obtained by the proposed convex program \eqref{eq:twolayerconvexprogram} and its approximation, respectively. Learned decision boundaries are also depicted. }\label{fig:minibatch_2d_anomaly}
\vskip -0.1in
\end{figure*}

We now present another numerical experiment on a two-dimensional dataset\footnote{In all the experiments, we use CVX \cite{cvx} and CVXPY \cite{cvxpy,cvxpy_rewriting} with the SDPT3 solver \cite{tutuncu2001sdpt3} to solve convex optimization problems.}, where we place a negative sample ($y=-1$) near the positive samples ($y=+1$) to have a more challenging loss landscape. In Figure \ref{fig:minibatch_2d_anomaly}, we observe that all the SGD realizations are stuck at local minima, therefore, achieve a significantly higher objective value compared to both Optimal and Approximate, which are based on convex optimization.

\begin{figure*}[h!]
\centering
\captionsetup[subfigure]{oneside,margin={1cm,0cm}}
	\begin{subfigure}[t]{0.45\textwidth}
	\centering
	\includegraphics[width=1.11\textwidth, height=0.8\textwidth]{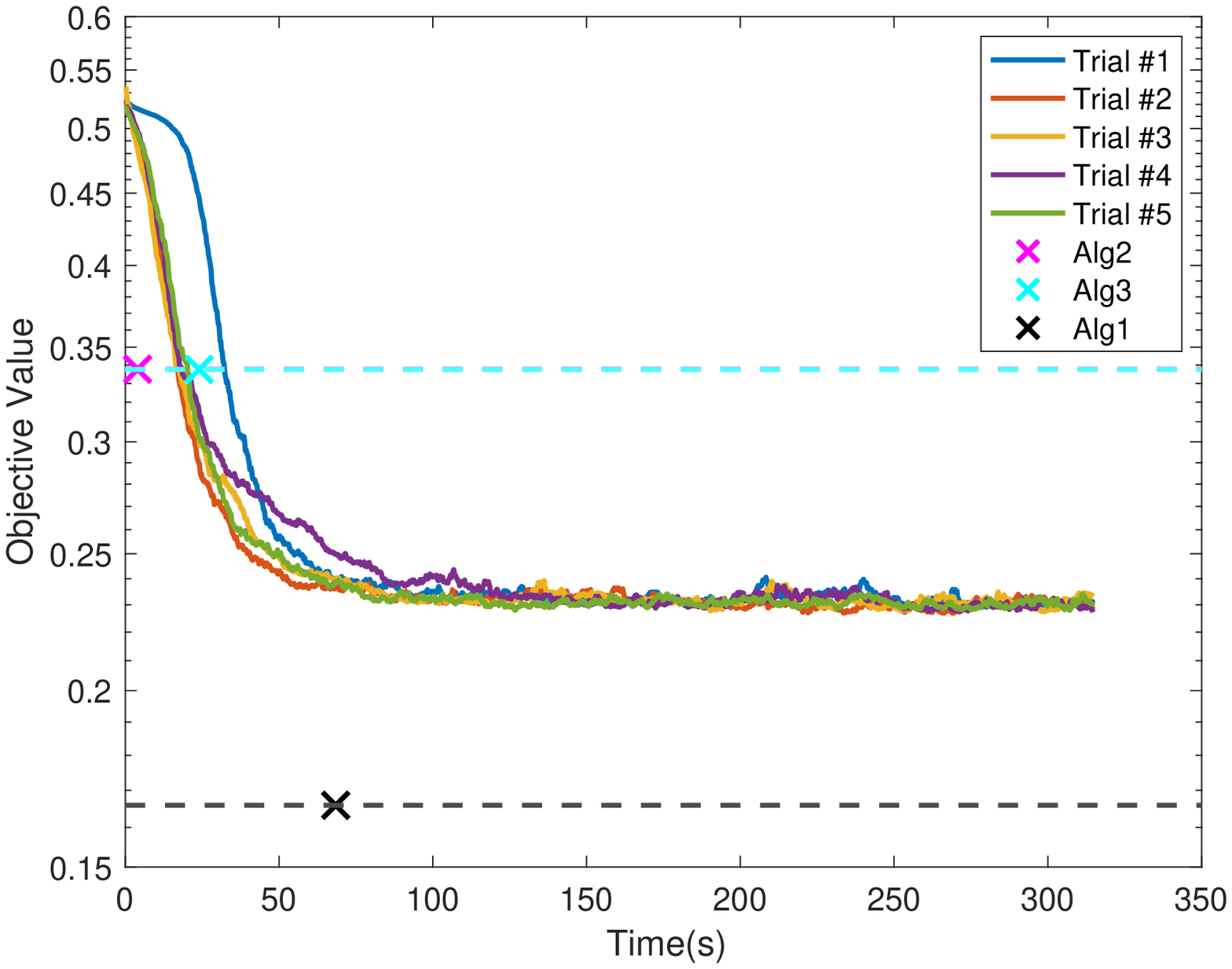}
	\caption{Training objective value\centering} \label{fig:boston_obj}
\end{subfigure} \hspace*{\fill}
	\begin{subfigure}[t]{0.45\textwidth}
	\centering
	\includegraphics[width=1.11\textwidth, height=0.8\textwidth]{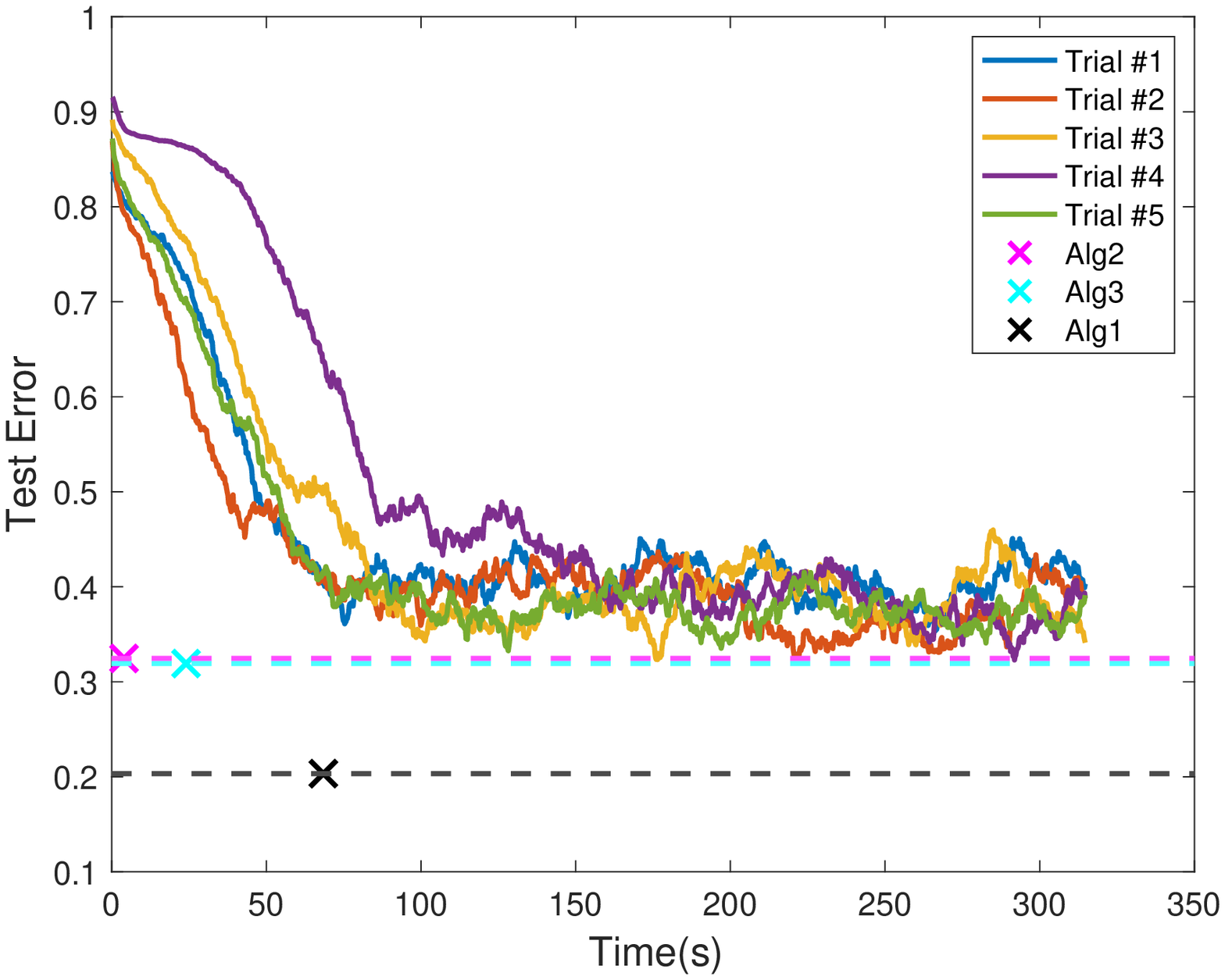}
	\caption{Test error\centering} \label{fig:boston_test}
\end{subfigure} \hspace*{\fill}
\caption{Training and test errors of the algorithms on the Boston Housing dataset ($n=400$ and $d=13$) where we run SGD independently in 5 initialization trials. For the convex program \eqref{eq:twolayerconvexprogram} approximations (Alg1, Alg2 and Alg3), crossed markers correspond to the total computation time of the convex optimization solver.}\label{fig:boston}
\vskip -0.1in
\end{figure*}

\begin{figure*}[h!]
\centering
\captionsetup[subfigure]{oneside,margin={1cm,0cm}}
	\begin{subfigure}[t]{0.45\textwidth}
	\centering
	\includegraphics[width=1.11\textwidth, height=0.8\textwidth]{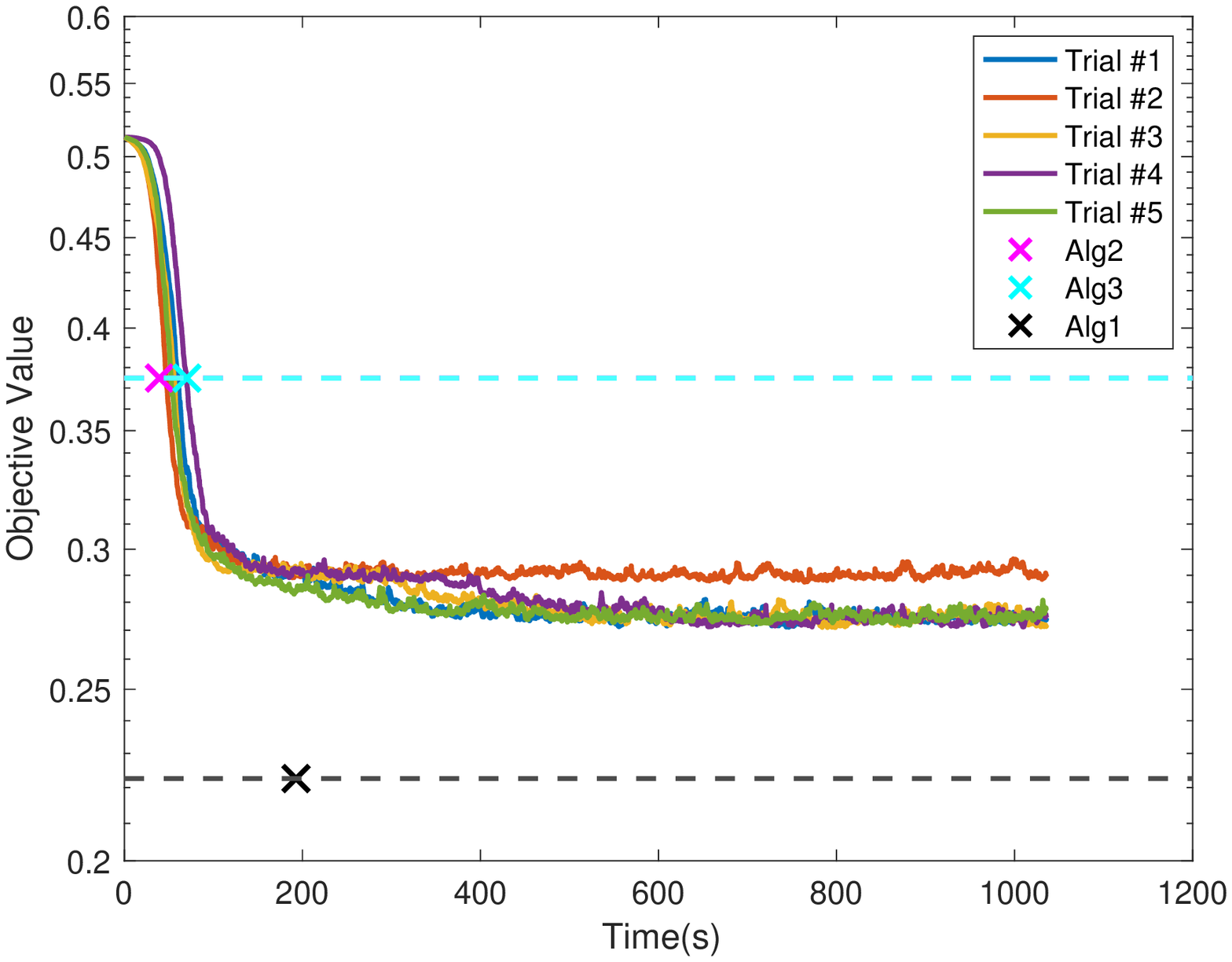}
	\caption{Training objective value\centering} \label{fig:kinematic_obj}
\end{subfigure} \hspace*{\fill}
	\begin{subfigure}[t]{0.45\textwidth}
	\centering
	\includegraphics[width=1.11\textwidth, height=0.8\textwidth]{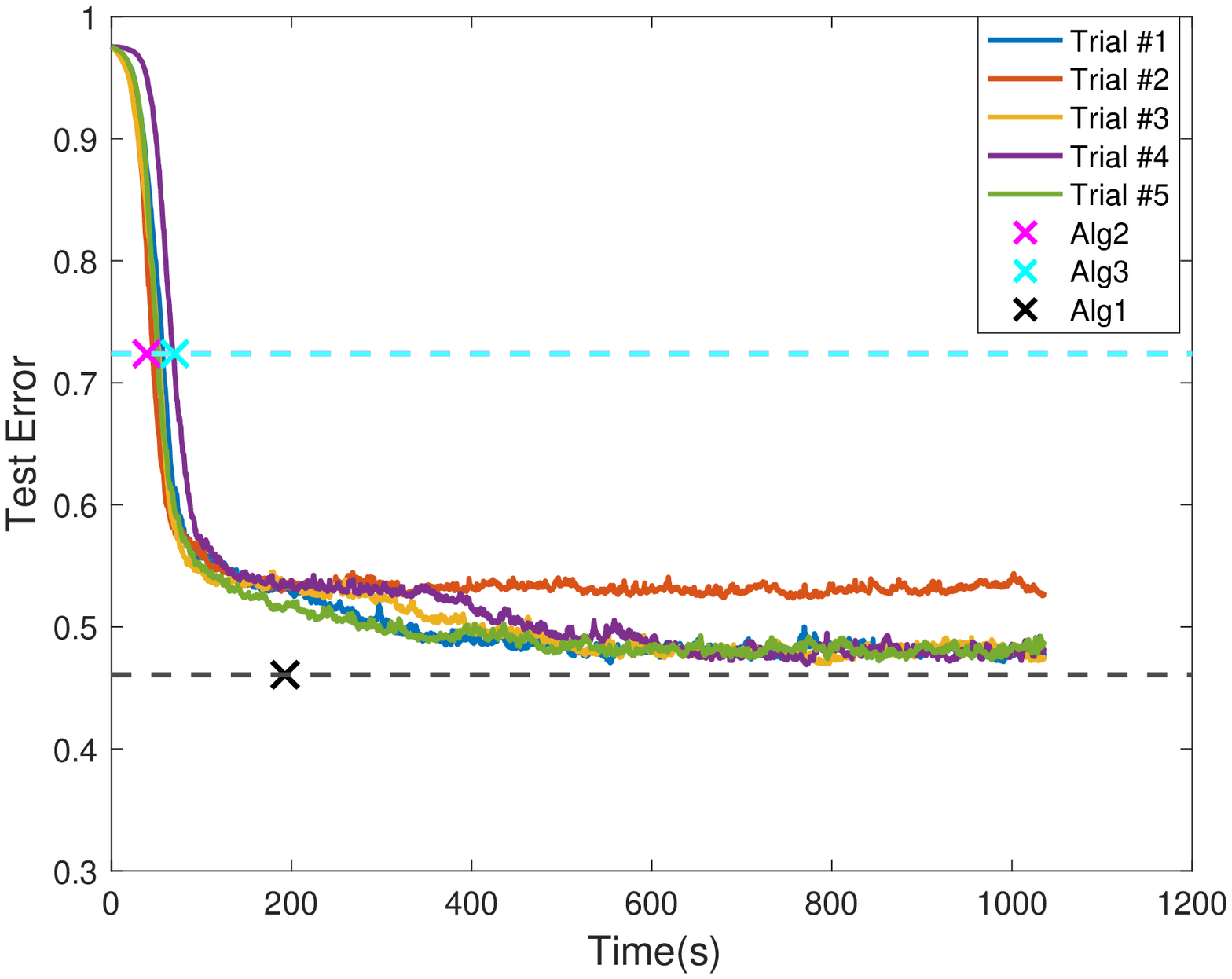}
	\caption{Test error\centering} \label{fig:kinematic_test}
\end{subfigure} \hspace*{\fill}
\caption{Performance comparison of the algorithms on the Kinematics dataset ($n=4000$ and $d=8$) where we run SGD independently in 5 initialization trials. For the convex program \eqref{eq:twolayerconvexprogram} approximations (Alg1, Alg2 and Alg3), crossed markers correspond to the total computation time of the convex optimization solver. }\label{fig:kinematic}
\vskip -0.1in
\end{figure*}

\begin{figure*}[h!]
\centering
\captionsetup[subfigure]{oneside,margin={1cm,0cm}}
	\begin{subfigure}[t]{0.45\textwidth}
	\centering
	\includegraphics[width=1.11\textwidth, height=0.8\textwidth]{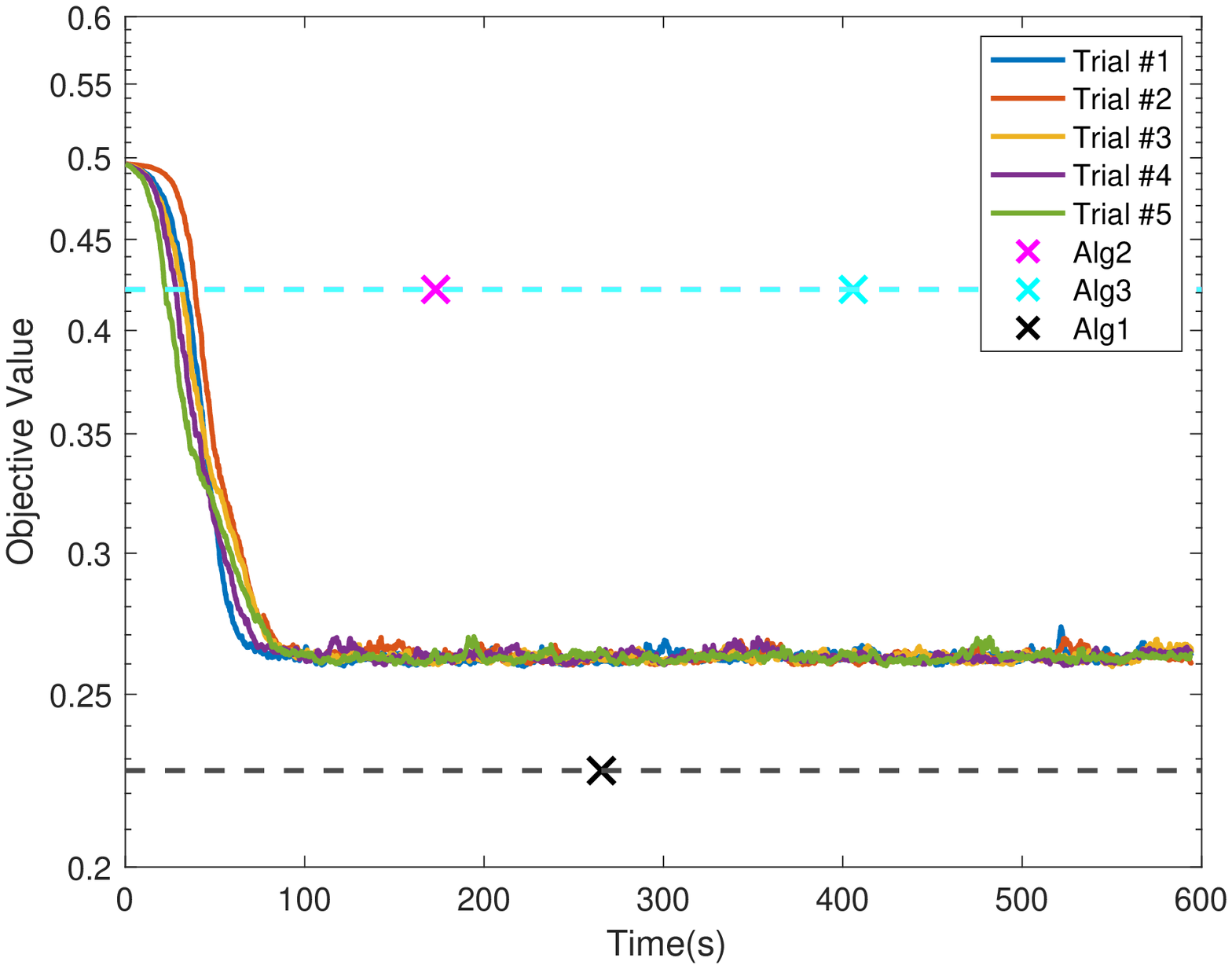}
	\caption{Training objective value\centering} \label{fig:bank_obj}
\end{subfigure} \hspace*{\fill}
	\begin{subfigure}[t]{0.45\textwidth}
	\centering
	\includegraphics[width=1.11\textwidth, height=0.8\textwidth]{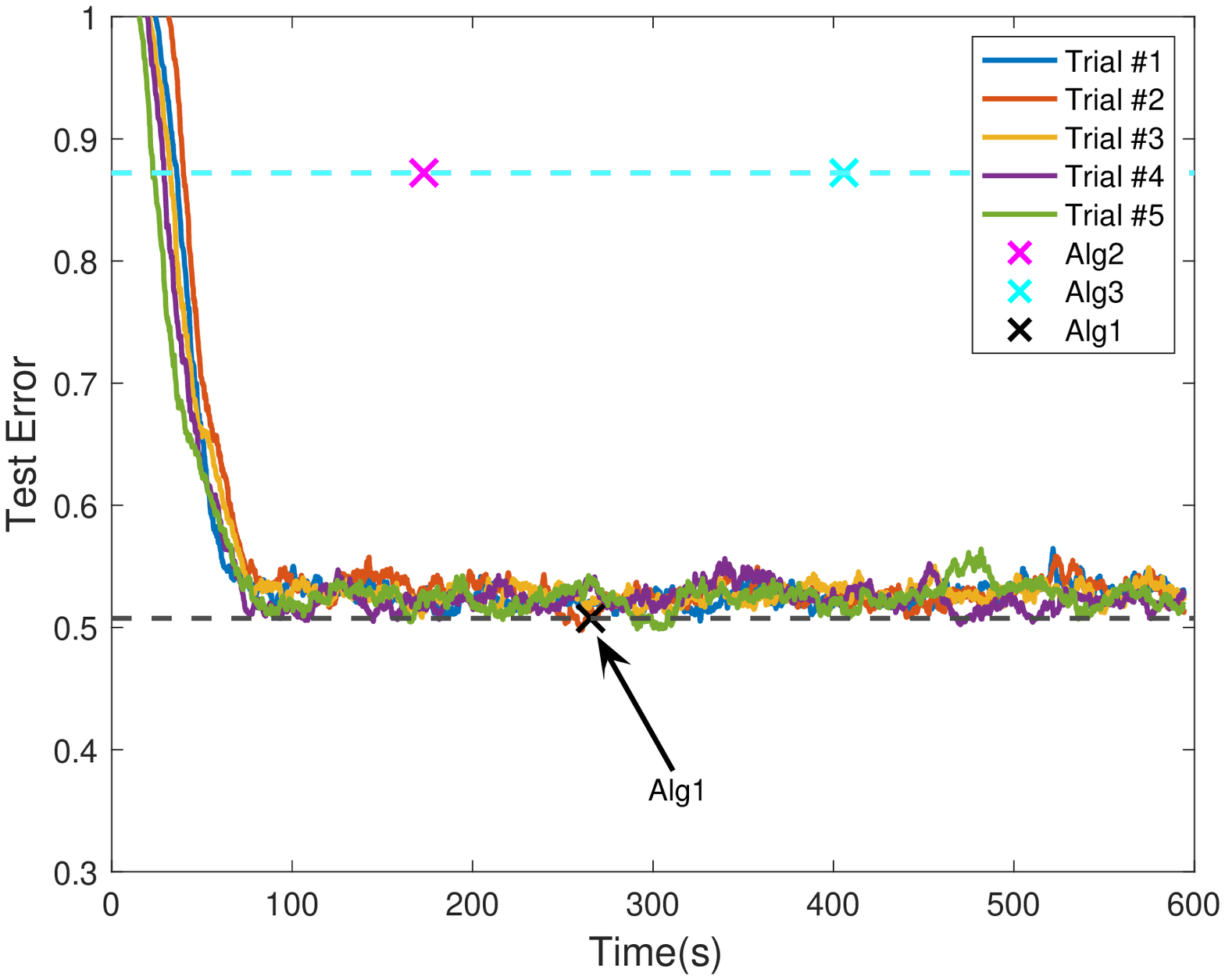}
	\caption{Test error\centering} \label{fig:bank_test}
\end{subfigure} \hspace*{\fill}
\caption{Performance comparison of the algorithms on the Bank dataset ($n=4000$ and $d=32$) where we run SGD independently in 5 initialization trials. For the convex program \eqref{eq:twolayerconvexprogram} approximations (Alg1, Alg2 and Alg3), crossed markers correspond to the total computation time of the convex optimization solver. }\label{fig:bank}
\vskip -0.1in
\end{figure*}

In addition to the classification datasets, we evaluate the performance of the algorithms on three regression datasets, i.e., the Boston Housing, Kinematics, and Bank datasets \cite{regression}. In Figure \ref{fig:boston}, we plot the objective value and the corresponding test error of $5$ independent initialization trials with respect to time in seconds, where we use squared loss and choose $n=400$, $d=13$, $m=12$, and batch size(bs) $25$. Similarly, we plot the objective values and test errors for the Kinematics and Bank datasets in Figure \ref{fig:kinematic} and \ref{fig:bank}, where $(n,d,m,\text{bs})=(4000,8,12,25)$ and $(n,d,m,\text{bs})=(4000,32,12,25)$, respectively. We observe that Alg1 achieves the lowest objective value and test error in both cases. 

\begin{figure*}[h!]
\centering
\captionsetup[subfigure]{oneside,margin={1cm,0cm}}
	\begin{subfigure}[t]{0.45\textwidth}
	\centering
	\includegraphics[width=1.11\textwidth, height=0.8\textwidth]{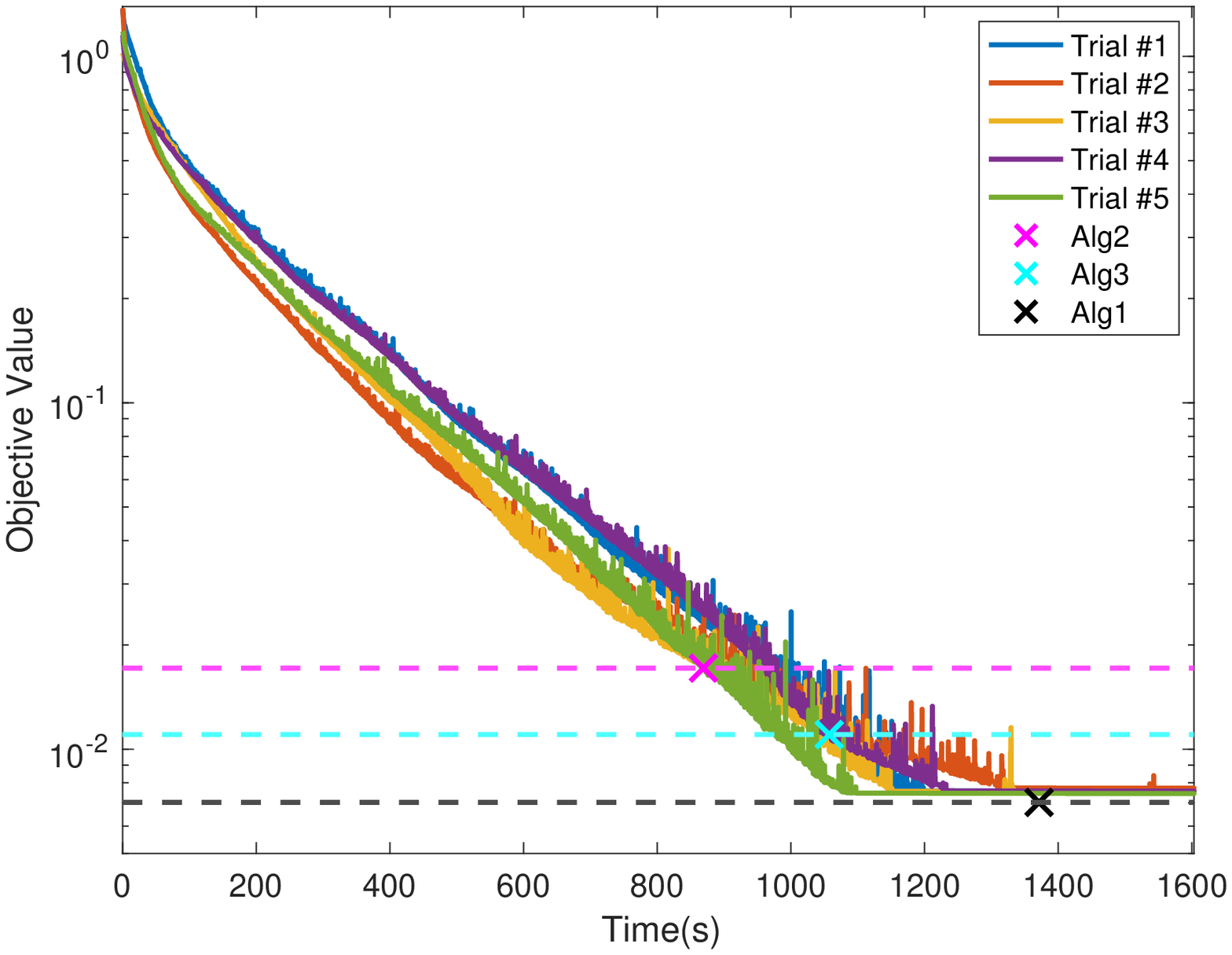}
	\caption{Training objective value\centering} \label{fig:relu_cnn_obj}
\end{subfigure} \hspace*{\fill}
	\begin{subfigure}[t]{0.45\textwidth}
	\centering
	\includegraphics[width=1.11\textwidth, height=0.8\textwidth]{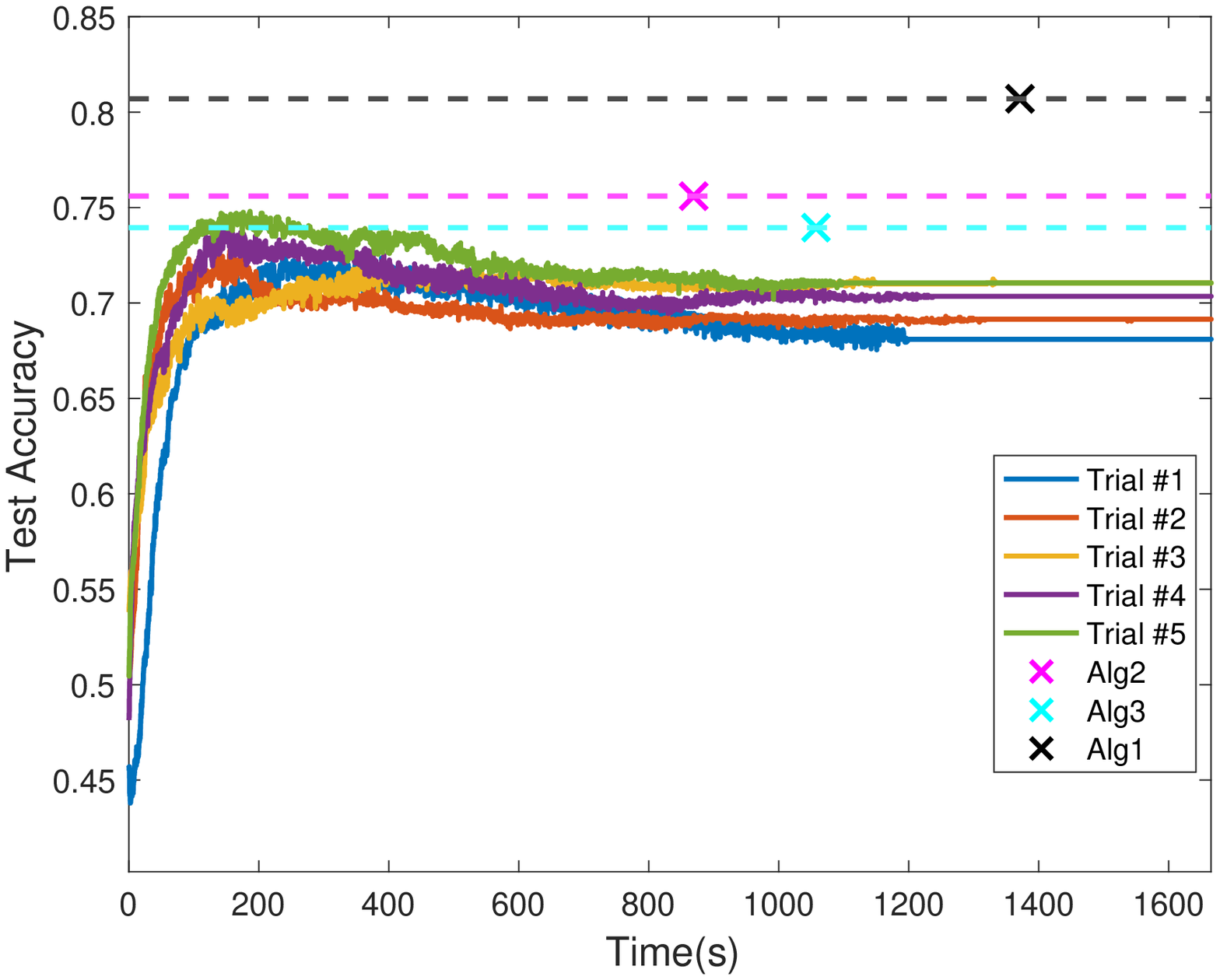}
	\caption{Test accuracy\centering} \label{fig:relu_cnn_test}
\end{subfigure} \hspace*{\fill}
\caption{Performance of the algorithms for two-layer CNN training on a subset of CIFAR-10 ($n=195$ and filter size $4\times 4 \times 3$) where we run SGD independently in 5 initialization trials. For the convex program \eqref{eq:twolayerconvexprogram} approximations (Alg1, Alg2 and Alg3), crossed markers correspond to the total computation time of the convex optimization solver. }\label{fig:relu_cnn}
\vskip -0.1in
\end{figure*}

\begin{figure*}[h!]
\centering
\captionsetup[subfigure]{oneside,margin={1cm,0cm}}
	\begin{subfigure}[t]{0.45\textwidth}
	\centering
	\includegraphics[width=1.11\textwidth, height=0.8\textwidth]{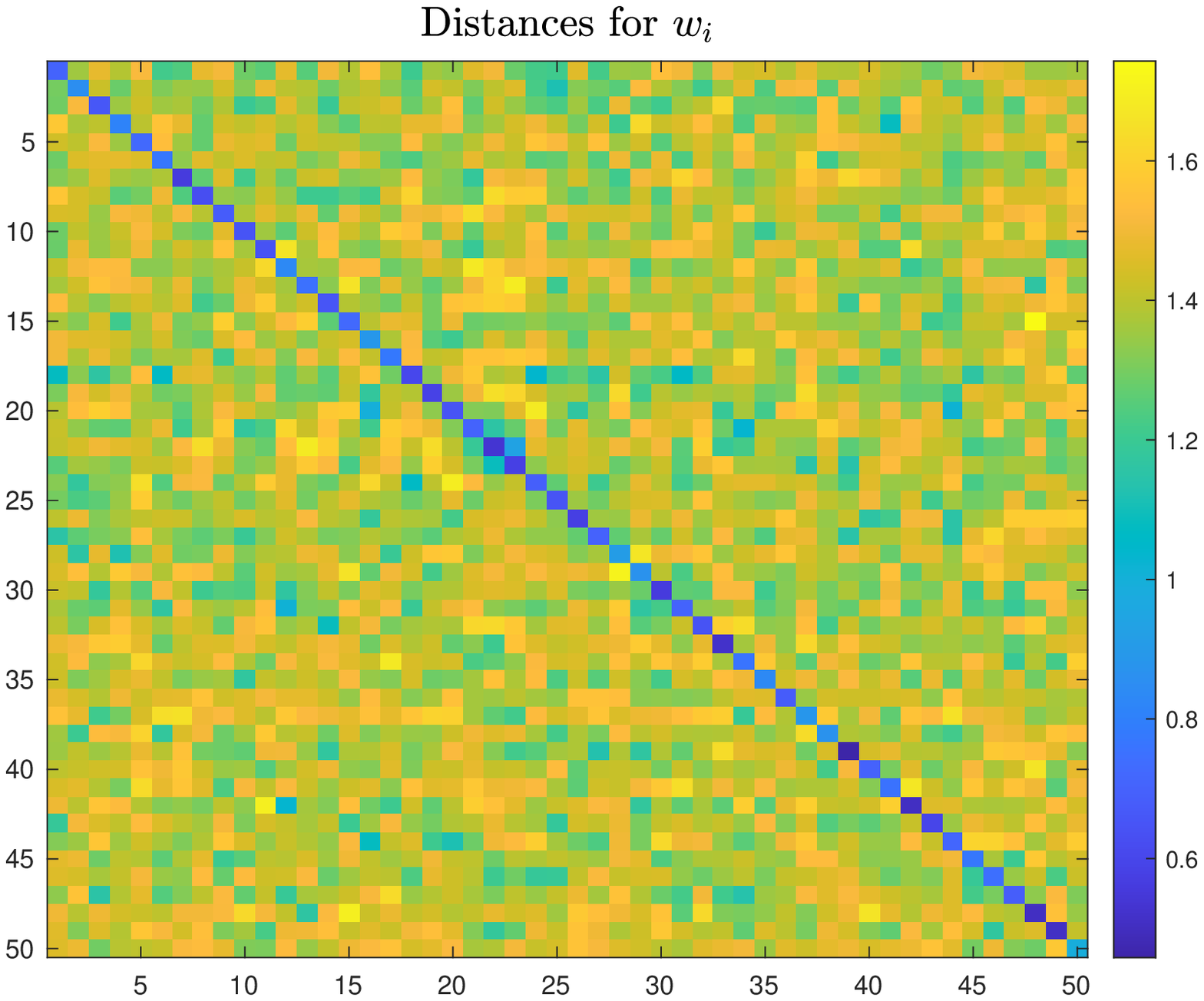}
	\caption{\centering} \label{fig:relu_cnn_dist1}
\end{subfigure} \hspace*{\fill}
	\begin{subfigure}[t]{0.45\textwidth}
	\centering
	\includegraphics[width=1.11\textwidth, height=0.8\textwidth]{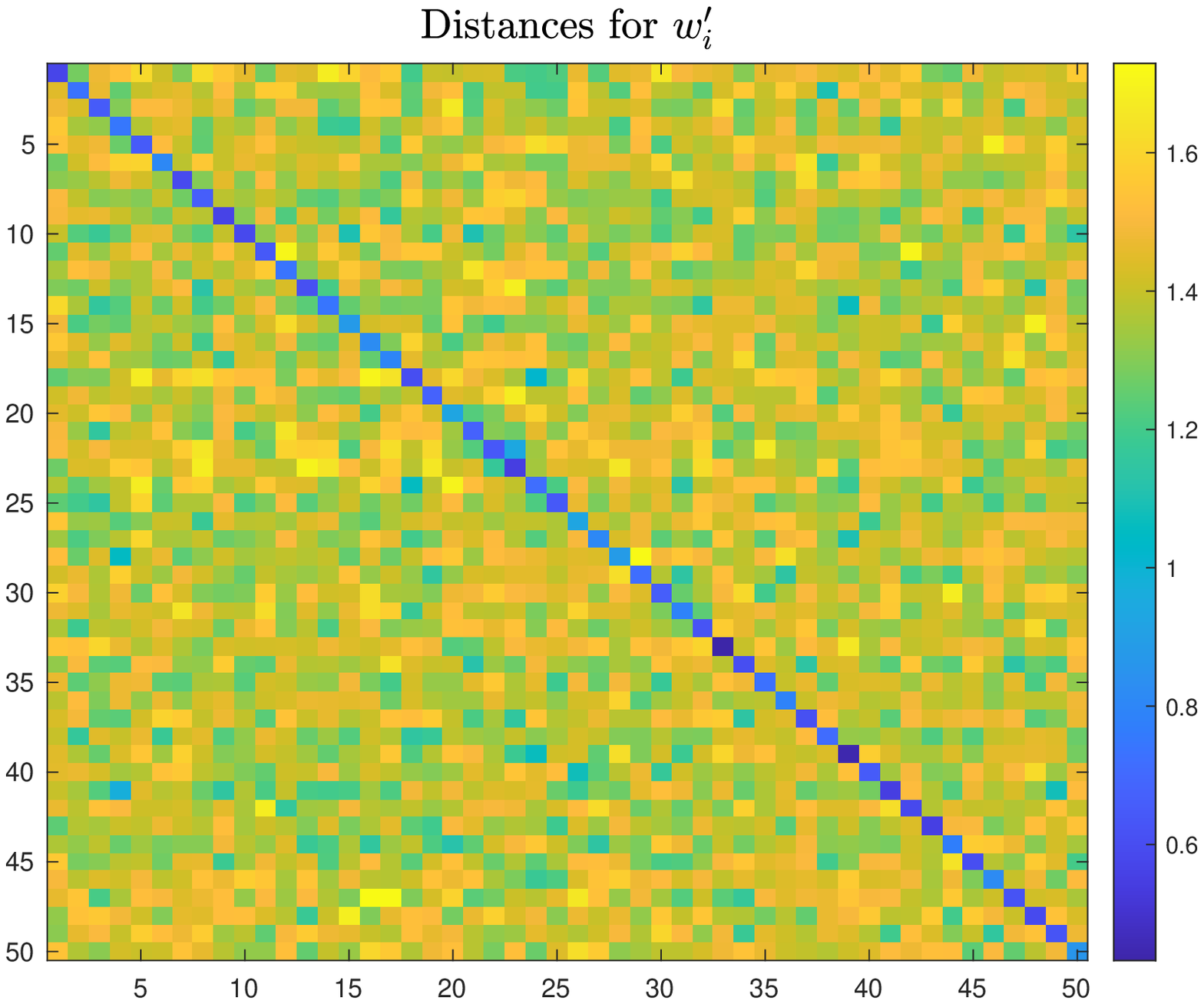}
	\caption{\centering} \label{fig:relu_cnn_dist2}
\end{subfigure} \hspace*{\fill}
\caption{Visualization of the distance (using the Euclidean norm of the difference) between the neurons found by GD and our convex program in Figure \ref{fig:relu_cnn}. The $ij^{th}$ entries of the distance plots are $\left\|\frac{w_i}{\|w_i\|_2}-\frac{\firstw_j}{\|\firstw_j\|_2}\right\|_2$ and $\left\|\frac{w_i^\prime}{\|w_i^\prime\|_2}-\frac{\firstw_j}{\|\firstw_j\|_2}\right\|_2$, respectively.}\label{fig:distance}
\vskip -0.1in
\end{figure*}

We also consider the training of a two-layer CNN architecture.
In Figure \ref{fig:relu_cnn}, we provide the binary classification performance of the algorithms on a subset of CIFAR-10, where we use hinge loss and choose $(n,d,m,\text{bs})=(195,3072,50,20)$, filter size $4\times 4 \times 3$, and stride $4$. This experiment also illustrates that Alg1 achieves lower objective value and higher test accuracy compared with the other methods including GD. We also emphasize that in this experiment, we use sign patterns of a clustered subset of patches, specifically $50$ clusters, as well as the GD patterns for Alg1. As depicted in Figure \ref{fig:distance}, the neurons that correspond to the sign patterns of GD matches with the neurons found by GD. Thus, the performance difference stems from the additional sign patterns found by clustering the patches.

In order to evaluate the computational complexity of the introduced approaches, in Table \ref{tab:training_time}, we provide the training time of each algorithm in the main paper. This data clearly shows that the introduced convex programs outperform GD while requiring significantly less training time.


\begin{table}[htb]
  \centering
\begin{small}
  \caption{Training time(in seconds), final objective value and test accuracy($\%$) of each algorithm in the main paper,where we use the CVX SDPT3 solver to optimize the convex programs.}
  \label{tab:training_time}
  \begin{tabular}{*{12}c}
    \toprule
    &\multicolumn{2}{c}{Figure \ref{fig:sgd_1d}}  & \multicolumn{3}{c}{Figure \ref{fig:minibatch_2d}}  & \multicolumn{4}{c}{Figure \ref{fig:relu_cifar}}  & \multicolumn{2}{c}{Figure \ref{fig:linear_cnn}}  \\ \cmidrule(l){2-3} \cmidrule(l){4-6} \cmidrule(l){7-10} \cmidrule(l){11-12}
& SGD  & Optimal &GD & Approx. & Optimal & GD & Alg1 & Alg2 & Alg3   &GD & L1-Convex   
    \\  \cmidrule(l){2-3} \cmidrule(l){4-6} \cmidrule(l){7-10}\cmidrule(l){11-12}
    \textbf{Time(s)}  & 420.663 & 1.225 &890.339 &1.498&117.858& 624.787&108.065&5.931&12.009& 65.365&1.404  \\
    \textbf{Train. Objective} &0.001&0.001& 0.0032&0.0028&0.0026&0.0042&0.0022&0.0032 &0.0032&0.804&0.803\\\
    \textbf{Test Accuracy($\%$)} & -& -&-&-&-&62.75&66.80&60.15&60.20&-&-\\
    \bottomrule
  \end{tabular}
  \end{small}
\end{table}

\subsection{Constructing hyperplane arrangements in polynomial time}\label{sec:hyperplane_arrangments_appendix}
We now consider the number of all distinct sign patterns $\mbox{sign}(Xz)$ for all possible choices $z\in \real^{d}$. Note that this number is the number of regions in a partition of $\real^d$ by hyperplanes passing through the origin, and are perpendicular to the rows of $\data$. We now show that the dimension $d$ can be replaced with $\mbox{rank}(\data)$ without loss of generality. Suppose that the data matrix $\data$ has rank $r$. We may express $\data = U\Sigma V^T$ using its Singular Value Decomposition in compact form, where $U\in\real^{n\times r}, \Sigma \in \real^{r\times r}, V^T \in \real^{r\times d}$. For any vector $z\in\real^d$ we have $\data z = U\Sigma V^Tz=Uz^\prime$  for some $z^\prime \in \real^r$.  Therefore, the number of distinct sign patterns $\mbox{sign}(Xz)$ for all possible $z\in \real^d$ is equal to the number of distinct sign patterns $\mbox{sign}(Uz^\prime)$ for all possible $z^\prime \in \real^r$.

Consider an arrangement of $n$ hyperplanes $\in \real^r$, where $n \ge r$. Let us denote the number of regions in this arrangement by $P_{n,r}$. In \cite{ojha2000enumeration,cover1965geometrical} it's shown that this number satisfies
\begin{align*}
    P_{n,r} \le 2\sum_{k=0}^{r-1}{n-1 \choose k}\,.
\end{align*}
For hyperplanes in general position, the above inequality is in fact an equality.
In \cite{edelsbrunner1986constructing}, the authors present an algorithm that enumerates all possible hyperplane arrangements $O(n^r)$ time, which can be used to construct the data for the convex program \eqref{eq:twolayerconvexprogram}.
%

\subsection{Equivalence of the $\ell_1$ penalized neural network training cost }\label{sec:equivalence_appendix}
In this section, we prove the equivalence between \eqref{eq:2layer_regularized_cost} and \eqref{eq:2layer_regularized_cost_l1}. 
\begin{lemma}[\cite{neyshabur_reg,infinite_width,ergen2020convex,ergen2020convex2,ergen2020convexdeep}] \label{lemma:scaling}
The following two problems are equivalent:
\begin{align*}
  \begin{split}
      &\min_{\{\firstw_j,\secondw_j\}_{j=1}^m} \frac{1}{2}\left\| \sum_{j=1}^m (\data \firstw_j )_+ \secondw_j -\labelvec \right\|_2^2+\frac{\beta}{2}\sum_{j=1}^m (\|\firstw_j\|_2^2 +\secondw_j^2 )
  \end{split}=\begin{split}
   \min_{\|u_j\|_2\le 1} \min_{\{\secondw_j\}_{j=1}^m} \frac{1}{2}\Big\| \sum_{j=1}^m (\data \firstw_j )_+ \secondw_j -\labelvec \Big\|_2^2 +\beta \sum_{j=1}^m \vert \secondw_j \vert\,
  \end{split}.
\end{align*}
\end{lemma}
[\textbf{Proof of Lemma \ref{lemma:scaling}}]
We can rescale the parameters as $\bar{\firstw}_{j}=\gamma_j\firstw_{j}$ and $\bar{\secondw}_{j}= \secondw_{j}/\gamma_j$, for any $\gamma_j>0$. Then, the output becomes
\begin{align*}
   \sum_{j=1}^m (\data \bar{\firstw}_{j})_+\bar{\secondw}_{j} =\sum_{j=1}^m ( \data \firstw_{j}\gamma_j)_+\frac{\secondw_{j}}{\gamma_j} =\sum_{j=1}^m (\data \firstw_{j})_+\secondw_{j} ,
\end{align*}
which proves that the scaling does not change the network output. In addition to this, we have the following basic inequality
\begin{align*}
   \frac{1}{2} \sum_{j=1}^m (\secondw_{j}^2+\| \firstw_{j}\|_2^2) \geq \sum_{j=1}^m (|\secondw_{j}| \text{ }\| \firstw_{j}\|_2),
\end{align*}
where the equality is achieved with the scaling choice $\gamma_j=\big(\frac{|\secondw_{j}|}{\| \firstw_{j}\|_2}\big)^{\frac{1}{2}}$ is used. Since the scaling operation does not change the right-hand side of the inequality, we can set $\|\firstw_{j} \|_2=1, \forall j$. Therefore, the right-hand side becomes $\| \secondw\|_1$.

Now, let us consider a modified version of the problem, where the unit norm equality constraint is relaxed as $\| \firstw_{j} \|_2 \leq 1$. Let us also assume that for a certain index $j$, we obtain  $\| \firstw_{j} \|_2 < 1$ with $\secondw_{j}\neq 0$ as an optimal solution. This shows that the unit norm inequality constraint is not active for $\firstw_{j}$, and hence removing the constraint for $\firstw_{j}$ will not change the optimal solution. However, when we remove the constraint, $\| \firstw_{j}\|_2 \rightarrow \infty$ reduces the objective value since it yields $\secondw_{j}=0$. Therefore, we have a contradiction, which proves that all the constraints that correspond to a nonzero $\secondw_{j}$ must be active for an optimal solution. This also shows that replacing $\|\firstw_j\|_2=1$ with $\| \firstw_{j} \|_2 \leq 1$ does not change the solution to the problem.


\subsection{Dual problem for \eqref{eq:2layer_regularized_cost_l1}}\label{sec:twolayer_dualform_appendix}
The following lemma proves the dual form of \eqref{eq:2layer_regularized_cost_l1}.
\begin{lemma}\label{lem:twolayer_dual}
The dual form of the following primal problem
\begin{align*}
     \min_{\|u_j\|_2\le 1} \min_{\{\secondw_j\}_{j=1}^m} \frac{1}{2}\Big\| \sum_{j=1}^m (\data \firstw_j )_+ \secondw_j -\labelvec \Big\|_2^2 +\beta \sum_{j=1}^m \vert \secondw_j \vert\,,
\end{align*}
is given by the following
\begin{align*}
     \min_{\|u_j\|_2\le 1} \max_{ \substack{v \in \real^n \,\mbox{\scriptsize s.t.} \\ \vert v^T(\data u_j)_+\vert \le \beta  }} -\frac{1}{2} \|y-v\|_2^2 + \frac{1}{2}\|y\|_2^2\,.
\end{align*}
\end{lemma}
[\textbf{Proof of Lemma \ref{lem:twolayer_dual}}]
Let us first reparametrize the primal problem as follows
\begin{align*}
     \min_{\|u_j\|_2\le 1} \min_{r,\{\secondw_j\}_{j=1}^m} \frac{1}{2}\| r\|_2^2 +\beta \sum_{j=1}^m \vert \secondw_j \vert\, \text{ s.t. }r=\sum_{j=1}^m (\data \firstw_j )_+ \secondw_j -\labelvec ,
\end{align*}
which has the following Lagrangian
\begin{align*}
    L(v,r,\firstw_j,\secondw_j)=\frac{1}{2}\| r\|_2^2 +\beta \sum_{j=1}^m \vert \secondw_j \vert+v^Tr +v^Ty -v^T\sum_{j=1}^m (\data \firstw_j )_+ \secondw_j.
\end{align*}
Then, minimizing over $r$ and $\alpha$ yields the proposed dual form.

%

\subsection{Dual problem for \eqref{eq:SDP} }\label{sec:sdp_appendix}
Let us first reparameterize the primal problem as follows
\begin{align*}
    &\max_{M,\dual} -\frac{1}{2}\|v-y\|_2^2+\frac{1}{2}\|y\|_2^2 \mbox{ s.t.  } \sigma_{\max}\left(M\right)\le \beta, \; M=[\data_1^T \dual\, ...\, \data_K^T \dual ].
\end{align*}
Then the Lagrangian is as follows
\begin{align*}
    L(\lambda,Z,M,v)&=-\frac{1}{2}\|v-y\|_2^2+\frac{1}{2}\|y\|_2^2+\lambda\left(\beta-\sigma_{\max}\left(M\right)\right)+\trace(Z^T M)-\trace(Z^T[\data_1^T \dual\, ...\, \data_K^T \dual ])\\
    &=-\frac{1}{2}\|v-y\|_2^2+\frac{1}{2}\|y\|_2^2+\lambda\left(\beta-\sigma_{\max}\left(M\right)\right)+\trace(Z^T M)-\dual^T \sum_{k=1}^K \data_kz_k,
\end{align*}
where $\lambda \geq 0$. Then maximizing over $M$ and $v$ yields the following dual form
\begin{align*}
    &\min_{z_k\in\real^d ,\forall k \in [K] } \frac{1}{2}\Big\|\sum_{k=1}^K \data_k z_k-y \Big\|^2_2+ \beta\Big \|[z_1,...,z_K] \Big \|_{*},
\end{align*}
where $\Big\|[z_1,...,z_K] \Big \|_{*}=\|Z\|_*=\sum_i \sigma_i(Z)$ is the $\ell_1$ norm of singular values, i.e., nuclear norm \cite{recht2010guaranteed}.  
%


\subsection{Dual problem for \eqref{eq:linear_cnn} }\label{sec:circular_linear_cnn_appendix}
Let us denote the eigenvalue decomposition of $U_j$ as $U_j=F D_j F^H$, where $F \in \mathbb{C}^{d \times d}$ is the Discrete Fourier Transform matrix and $D_j\in \mathbb{C}^{d \times d}$ is a diagonal matrix. Then, applying the scaling in Lemma \ref{lemma:scaling} and then taking the dual as in Lemma \ref{lem:twolayer_dual} yields
\begin{align*}
    \max_\dual -\frac{1}{2}\|\dual-y\|_2^2 +\frac{1}{2}\| y\|_2^2 \text{ s.t. } \|\dual^T \data F D F^H \|_2 \leq \beta,\; \forall D: \|D \|_F^2 \leq d,
\end{align*}
which can be equivalently written as
\begin{align*}
    \max_\dual -\frac{1}{2}\|\dual-y\|_2^2 +\frac{1}{2}\| y\|_2^2 \text{ s.t. } \|\dual^T \tilde{\data} D \|_2 \leq \beta,\; \forall D: \|D \|_F^2 \leq d .
\end{align*}
Since $D$ is diagonal, $\|D \|_F^2 \leq d$ is equivalent to $\sum_{i=1}^d D_{ii}^2 \leq 1$. Therefore, the problem above can be further simplified as
\begin{align*}
    \max_\dual -\frac{1}{2}\|\dual-y\|_2^2 +\frac{1}{2}\| y\|_2^2 \text{ s.t. } \|\dual^T \tilde{\data}\|_{\infty} \leq \frac{\beta}{\sqrt{d}}\;  .
\end{align*}
Then, taking the dual of this problem gives the following
\begin{align*}
    \min_{z \in \mathbb{C}^d} \frac{1}{2}\Big \|\tilde{\data} z -y \Big\|^2_2+ \frac{\beta}{\sqrt{d}} \|z\|_1.
\end{align*}


\subsection{Dual problem for vector output two-layer linear convolutional networks}\label{sec:vectoroutput_appendix}
Vector version of the two-layer linear convolutional network training problem has the following dual
\begin{align*}
    &\max_{\dualmat} \, \mathrm{trace} \, \dualmat^T \vec{Y} \mbox{ s.t.  } \max_{\|\firstw\|_2\le 1}\, \sum_{k} \| \dualmat^T \data_k \firstw \|_2^2 \le 1.
\end{align*}
Similarly, extreme points are the maximal eigenvectors of $\sum_k \data_k^T \dualmat \dualmat^T \data_k$ When $\dualmat=\vec{Y}$, and one-hot encoding is used, these are the right singular vectors of the matrix
$[\data^T_{1,c}\, \data^T_{2,c}\, ...\, \data^T_{K,c}]^T$ whose rows contain all the patch vectors for class $c$.

\subsection{Semi-infinite strong duality \label{appendix_semi_infinite_duality}}
Note that the semi-infinite problem \eqref{eq:2layer_regularized_cost_innerdual} is convex. We first show that the optimal value is finite. For $\beta>0$, it is clear that $v=0$ is strictly feasible, and achieves $0$ objective value. Note that the optimal value $p^*$ satisfies $p^*\le \|y\|_2^2$ since this value is achieved when all the parameters are zero.
Consequently, Theorem 2.2 of \cite{shapiro2009semi} implies that strong duality holds, i.e., $p^*=d^*_{\infty}$, if the solution set of the semi-infinite problem in \eqref{eq:2layer_regularized_cost_innerdual} is nonempty and bounded. Next, we note that the solution set of \eqref{eq:2layer_regularized_cost_innerdual} is the Euclidean projection of $y$ onto the polar set $(\rectset \cup -\rectset)^\circ$ which is a convex, closed and bounded set since $\relu{\data\firstw}$ can be expressed as the union of finitely many convex closed and bounded sets.
\qed

\subsection{Semi-infinite strong gauge duality \label{appendix_semi_infinite_duality_gauge}}
Now we prove strong duality for \eqref{eq:supportfun}. We invoke the semi-infinite optimality conditions for the dual \eqref{eq:supportfun}, in particular we apply Theorem 7.2 of \cite{semiinfinite_goberna} and use the standard notation therein. We first define the set
\begin{align*}
    \mathbf{K}=\mathbf{cone}\left\{ \left( \begin{array}{c}s\, \relu{\data \firstw} \\ 1 \end{array}  \right), \firstw \in \ball_2, s\in\{-1,+1\}; \left(\begin{array}{c} \vec{0} \\ -1\end{array}\right) \right\}\,.
\end{align*}
Note that $\mathbf{K}$ is the union of finitely many convex closed sets, since $\relu{\data\firstw}$ can be expressed as the union of finitely many convex closed sets. Therefore the set $\mathbf{K}$ is closed. By Theorem 5.3 \cite{semiinfinite_goberna}, this implies that the set of constraints in \eqref{eq:dual} forms a Farkas-Minkowski system. By Theorem 8.4 of \cite{semiinfinite_goberna}, primal and dual values are equal, given that the system is consistent. Moreover, the system is discretizable, i.e., there exists a sequence of problems with finitely many constraints whose optimal values approach to the optimal value of \eqref{eq:dual}.
\qed

\subsection{Neural Gauge function and equivalence to minimum norm networks}\label{sec:gauge_appendix}
Consider the gauge function
\begin{align*}
    p^g=&\min_{r\ge0}\, r \mbox{ s.t. } ry \in \mbox{conv}(\rectset\cup-\rectset)
\end{align*}
and its dual representation in terms of the support function of the polar of $\mbox{conv}(\rectset\cup-\rectset)$
\begin{align*}
    d^g=\max_{v} v^T y  \mbox{ s.t. } v\in (\rectset\cup-\rectset)^\circ.
\end{align*}
Since the set $\rectset\cup-\rectset$ is a closed convex set that contains the origin, we have $p^g=d^g$ \cite{Rockafellar} and $\left(\mbox{conv}(\rectset\cup-\rectset)\right)^\circ = (\rectset\cup-\rectset)^\circ$. The result in Section \ref{appendix_semi_infinite_duality} implies that the above value is equal to the semi-infinite dual value, i.e., $p^d=p^g_{\infty}$, where 
\begin{align*}
    p^g_{\infty} := \min_{\mu}\, \|\mu\|_{TV}\mbox{ s.t. } \int_{u \in \ball_2} (\data u)_+d\mu(u)=y\,.
\end{align*}
By Caratheodory's theorem, there exists optimal solutions the above problem consisting of $m^*$ Dirac deltas \cite{Rockafellar,rosset2007}, and therefore
\begin{align*}
    p^g_{\infty} = \min_{{u_j \in \ball_2},j\in[m^*]}\, \sum_{j=1}^{m^*} \vert\alpha_j\vert \mbox{ s.t. } \sum_{j=1}^{m^*} (\data u_j)_+d\alpha_j=y\,,
\end{align*}
where we define $m^*$ as the number of Dirac delta's in the optimal solution to $p^g_{\infty}$. If the optimizer is non-unique, we define $m^*$ as the minimum cardinality solution among the set of optimal solutions.
Now consider the non-convex problem
\begin{align*}
    &\min_{\{u_j,\alpha_j\}_{j=1}^m} \|\alpha\|_1 \mbox{ s.t. } \sum_{j=1}^m(Xu_j)_+ \alpha_j =y,\;\|u_j\|_2\le 1\,.
\end{align*}
Using the standard parameterization for $\ell_1$ norm we get
\begin{align*}
    &\min_{\{u_j\}_{j=1}^m,s\ge0,t\ge0} \sum_{j=1}^m (t_j + s_j) \mbox{ s.t. } \sum_{j=1}^m(Xu_j)_+ t_j - (Xu_j)_+s_j=y,\; \|u_j\|_2\le 1\,,\forall j.
\end{align*}
Introducing a slack variable $r\in\reals_+$, an equivalent representation can be written as
\begin{align*}
    &\min_{\{u_j\}_{j=1}^m,s\ge0,t\ge0,r\ge0}\, r \mbox{ s.t. } \sum_{j=1}^m(Xu_j)_+ t_j - (Xu_j)_+s_j=y,\; \sum_{j=1}^m (t_j + s_j )= r,\;\|u_j\|_2\le 1\,,\forall j.
\end{align*}
Note that $r>0$ as long as $y\neq 0$. Rescaling variables by letting $t_j^\prime = t_j/r$, $s_j^\prime = s_j/r$ in the above program, we obtain
%
\begin{align}
    &\min_{\{u_j\}_{j=1}^m,s^\prime\ge0,t^\prime\ge0,r\ge0}\, r \label{eq:gaugenonconvex} \mbox{ s.t. } \sum_{j=1}^m \left((Xu_j)_+ t^\prime_j - (Xu_j)_+s^\prime_j\right)=ry,\; \sum_{j=1}^m (t_j^\prime+ s_j^\prime) = 1,\;\|u_j\|_2\le 1\,,\forall j \nonumber\,.
\end{align}
Suppose that $m\ge m^*$. It holds that
\begin{align}
    \exists s^\prime,t^\prime\ge0\, ,\{u_j\}_{j=1}^m\, \mbox{ s.t. } \sum_{j=1}^m (t_j^\prime+ s_j^\prime) = 1,\,\|u_j\|_2\le 1,\,\forall j,\,\sum_{j=1}^m (Xu_j)t^\prime_j - (Xu_j)_+s^\prime_j=ry \,\, \iff ry \in \mbox{conv}(\rectset\cup-\rectset).
\end{align}
We conclude that the optimal value of \eqref{eq:gaugenonconvex} is identical to the gauge function $p_g$.
\subsection{Alternative proof of the semi-infinite strong duality \label{appendix_semi_infinite_duality2}}
It holds that $p^* \ge d^*$ by weak duality in \eqref{eq:2layer_regularized_cost_innerdual}.
Theorem \ref{thm:mainconvex} proves that the objective value of \eqref{eq:dual} is identical to the value of \eqref{eq:2layer_regularized_cost} as long as $m\ge m^*$. Therefore we have $p^*=d^*$.
\qed

\subsection{Finite dimensional strong duality results for Theorem \ref{thm:mainconvex}}
\label{sec:finitestrongdualconst}
\begin{lemma}
\label{lem:finitestrongdualconst}
Suppose $D(S)$, $D(S^c)$ are fixed diagonal matrices as described earlier, and $\data$ is a fixed matrices. The dual of the convex optimization problem
\begin{align*}
&\max_{\substack{u \in \real^d \\ \|u\|_2\le 1 \\ D(S)\data u\ge 0 \\ D(S^c)\data u\le 0 }}
v^TD(S) \data u 
\end{align*}
is given by
\begin{align*}
&\qquad \min_{\substack{\alpha,\beta \in \real^{n } \\ \alpha,\beta \ge 0}}
\| \data^T D(S) \big(v + \alpha + \beta \big)-\data^T \beta \|_2\,
\end{align*}
and strong duality holds.
\end{lemma}
Note that the linear inequality constraints specify valid hyperplane arrangements. Then there exists strictly feasible points in the constraints of the maximization problem. Standard finite second order cone programming duality implies that strong duality holds \cite{boyd_convex} and the dual is as specified.
\qed

\subsection{General loss functions}\label{sec:generalloss_appendix}
In this section, we extend our derivations to arbitrary convex loss functions.

Consider minimizing the sum of the squared loss objective and squared $\ell_2$-norm of all parameters
\begin{align}
    p^*:=\min_{\{\secondw_j,\firstw_j\}_{j=1}^m} \ell\left( \sum_{j=1}^m (Xu_j)_+\alpha_j, y\right) +\frac{\beta}{2} \sum_{j=1}^m (\|\firstw_j\|_2^2+\secondw_j^2)\,,
    \label{eq:2layer_regularized_cost_general}
\end{align}
where $\ell(\cdot,y)$ is a convex loss function. Then, consider the following finite dimensional convex optimization problem
\begin{align}
&\min_{\{v_i,w_i\}_{i=1}^P}\, \ell\left( \sum_{i=1}^P D_iX(v_i-w_i) , y\right) +\beta\sum_{i=1}^P \left(\|v_i\|_2+\|w_i\|_2\right)   \mbox{ s.t. }  
(2D_i-I)Xv_i\ge0,~ (2D_i-I) Xw_i\ge 0.\, \forall i \in [P],\label{eq:twolayerconvexprogram_general}   
\end{align}
 Let us define $m^*=\sum_{\tiny i: v_i^*\neq 0}^P 1+\sum_{\tiny i:  w^*_i\neq 0}^P 1$, where $\{v^*_i,w^*_i\}_{i=1}^P$ are optimal in \eqref{eq:twolayerconvexprogram_general}.
\begin{theorem}
\label{thm:mainconvex_general}
The convex program \eqref{eq:twolayerconvexprogram_general} and the non-convex problem \eqref{eq:2layer_regularized_cost_general} where $m\ge m^*$ has identical optimal values. Moreover, an optimal solution to \eqref{eq:2layer_regularized_cost_general} can be constructed from an optimal solution to \eqref{eq:twolayerconvexprogram_general} as follows
\eqref{eq:twolayerconvexprogram} as follows
\begin{align*}
    (u^*_{j_{1i}},\alpha^*_{j_{1i}}) =  \left(\frac{v^*_i}{\sqrt{\|v^*_i\|_2}}, \sqrt{\|v^*_i\|_2}\right) \,\quad \quad \mbox{  if  } \quad v_i^*\neq 0  \\
    (u^*_{j_{2i}},\alpha^*_{j_{2i}}) =   \left(\frac{w^*_i}{\sqrt{\|w^*_i\|_2}}, -\sqrt{\|w^*_i\|_2}\right) \quad \mbox{  if  } \quad w_i^*\neq 0\,, 
\end{align*}
where $v_i^*,w_i^*$ are the optimal solutions to \eqref{eq:twolayerconvexprogram_general}.
\end{theorem}
[\textbf{Proof of Theorem \ref{thm:mainconvex_general}}]
The proof parallels the proof of the main result section and Theorem \ref{theo:twolayer_generic_dual}. We note that dual constraint set remains the same, and analogous strong duality results apply as we show next.

We also show that our dual characterization holds for arbitrary convex loss functions.
\begin{align}
    \min_{\{\firstw_j,\secondw_j\}_{j=1}^m} \ell \left(\sum_{j=1}^m(\data \firstw_j)_+\secondw_j,\vec{y}\right) + \beta \| \secondw\|_1 \text{ s.t. } \|\firstw_j\|_2\leq 1, \; \forall j,\label{eq:twolayer_general_loss}
\end{align}
where $\ell(\cdot,y)$ is a convex loss function. 
\begin{theorem}\label{theo:twolayer_generic_dual}
The dual of \eqref{eq:twolayer_general_loss} is given by
\begin{align*}
    &\max_{\dual}  - \ell^*(\dual)\mbox{ s.t. } \vert v^T(\data u)_+\vert \le \beta,\; \forall u \in \ball_2 \,,
\end{align*}
where $\ell^*$ is the Fenchel conjugate function defined as
\begin{align*}
\ell^*(\dual) = \max_{z} \vec{z}^T \dual - \ell(z,y)\,.
\end{align*}
\end{theorem}
[\textbf{Proof of Theorem \ref{theo:twolayer_generic_dual}}]
The proof follows from classical Fenchel duality \cite{boyd_convex}. We first describe \eqref{eq:twolayer_general_loss} in an equivalent form as follows
\begin{align*}
    \min_{z,\{\firstw_j,\secondw_j\}_{j=1}^m} \ell(\vec{z},y) + \beta \| \secondw \|_1 \text{ s.t. } \vec{z}=\sum_{j=1}^m(\data \firstw_j)_+\secondw_j, \; \| \firstw_{j} \|_2\leq1 , \forall j.
\end{align*}
Then the dual function is
\begin{align*}
    g(\dual)= \min_{z,\{\firstw_j,\secondw_j\}_{j=1}^m} \ell(z,y)- \dual^T z+ \dual^T \sum_{j=1}^m(\data\firstw_j)_+\secondw_j+ \beta \|\secondw\|_1\text{ s.t. } \| \firstw_j \|_2\leq1 , \forall j.
\end{align*}
Therefore, using the classical Fenchel duality \cite{boyd_convex} yields the claimed dual form.

\end{document}